\documentclass[11pt]{article}

  \usepackage{acl}
  \usepackage{times}
  \usepackage{latexsym}
  \usepackage[T1]{fontenc}
  \usepackage[utf8]{inputenc}
  \usepackage{microtype}
  \usepackage{inconsolata}
  \usepackage{graphicx}
  \usepackage{booktabs}
  \usepackage{amssymb} 
  \usepackage{enumitem}
  \usepackage{stfloats} % <--- Added to allow [b] placement for wide two-column figures
  \usepackage{amsmath}
  \usepackage{multirow}
  \usepackage{subcaption}% --- Packages and Definitions for Figure 1 (TikZ) ---
  \usepackage{tikz}
  \usetikzlibrary{positioning, calc}
  \usepackage{pgfplots}
  \usepgfplotslibrary{fillbetween}
  \pgfplotsset{compat=1.18}
  \usepackage{xcolor}
  \usepackage{tcolorbox}
  \tcbuselibrary{breakable,skins}

  \newtcolorbox{promptbox}[1][]{
      enhanced,
      breakable,
      colback=gray!4,
      colframe=black!40,
      boxrule=0.4pt,
      arc=1mm,
      left=2mm,
      right=2mm,
      top=1mm,
      bottom=1mm,
      fonttitle=\bfseries,
      title=#1
  }

  \definecolor{chain}{RGB}{46,117,182}
  \definecolor{cascade}{RGB}{192,57,43}
  \definecolor{conflict}{RGB}{39,174,96}
  \definecolor{counter}{RGB}{142,68,173}  
  \definecolor{temporal}{RGB}{211,84,0}
  \definecolor{baseline}{RGB}{127,140,141}
  \definecolor{logcolor}{RGB}{60,60,60}
  \definecolor{qbg}{RGB}{245,247,255}
  \definecolor{abg}{RGB}{245,255,248}
  \definecolor{dotcolor}{RGB}{175,175,175}

  \usepackage{hyperref}

  \title{RECON: Benchmarking Agent Memory for Compositional Reasoning over Long Contexts}

  \author{
    Mihir Shriniwas Arya \\
    Department of Computer Science and Engineering \\
    RV College of Engineering \\
    \texttt{mihirsarya.cy23@rvce.edu.in}
  }

\begin{document}
  \maketitle

  \begin{abstract}
  Large language models and LLM-based agents are widely used as personal chat assistants, enterprise copilots, and autonomous workflow agents. In all these applications, memory (the ability to retain, access, and reason over information accumulated over long contexts and multiple interactions) plays a crucial role in determining the reliability of any agent. We introduce RECON (Reasoning over Extended Contexts with Obfuscated Narratives), a benchmark for evaluating compositional reasoning over long contexts. RECON spans 24 case files across three domains (criminal, medical, and financial), each ranging from 50k to 100k tokens, and tests agents on \textbf{six memory-intensive tasks}: reconstructing multi-hop evidence chains, propagating cascading invalidations, resolving source conflicts, counterfactual reasoning, satisfying temporal constraints, and temporal fact retrieval. Recent memory benchmarks evaluate whether agents can retrieve scattered facts or detect if a fact has changed whereas RECON evaluates what happens after the change, whether agents can trace which downstream conclusions are affected, which survive through independent support, and how alternative timelines would have unfolded. Our evaluation reveals substantial limitations across current architectures: even the strongest non-Oracle system reaches only $22.4\%$ Accuracy, with retrieval and reasoning each surfacing as challenges. Our benchmark and code are available at \url{https://anonymous.4open.science/r/RECON-Bench}.

  \end{abstract}

  \begin{figure*}[t]
      \centering
      \includegraphics[width=\textwidth]{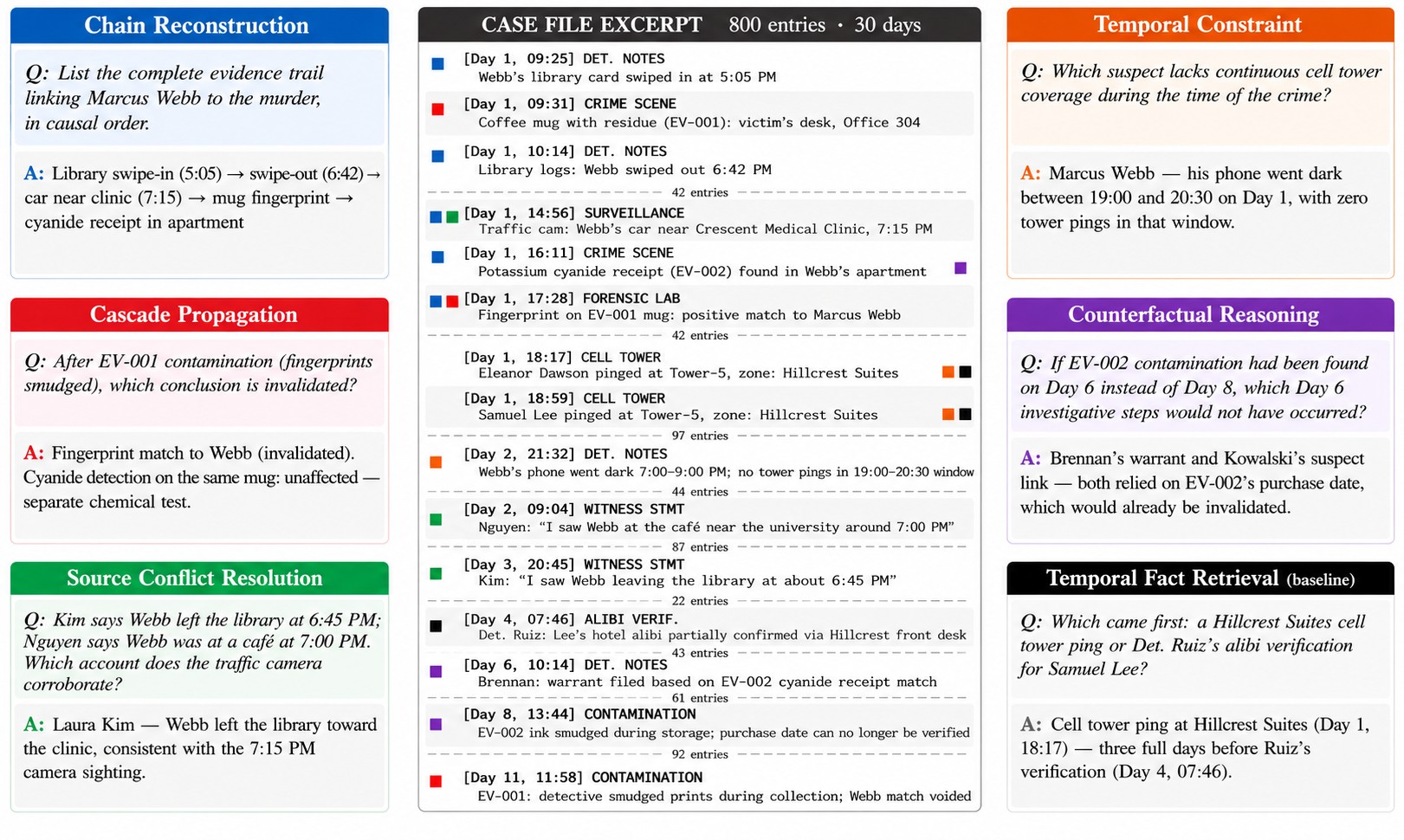}
      \caption{The six memory-intensive tasks evaluated by RECON.}
      \label{fig:overview}
  \end{figure*}

  \section{Introduction}

  Large Language Model based agents are tasked with increasingly high-stakes work. Coding assistants such as Claude Code and GitHub Copilot must track context across entire codebases \citep{ding2023crosscodeeval, liu2023repobench}, conversational agents such as ChatGPT and Claude must retain user preferences and instructions that change over time \citep{wu2024longmemeval, du2024perltqa}, and enterprise copilots must reason over evolving documents in clinical, legal, and financial workflows \citep{fleming2024medalign}. Without robust memory, an agent that forgets an earlier fact was revised, whether a lab result updated on Day 9, a witness statement contradicted on Day 5, or a flagged transaction reversed, hallucinates stale facts, misses critical updates, and loses track of how earlier evidence connects to later conclusions. Memory quality directly determines the quality of everything built on top of it.

  The importance of memory has driven a wave of memory architectures such as \textbf{Mem0} \citep{chhikara2025mem0}, \textbf{Zep} \citep{rasmussen2025zep}, \textbf{MemGPT} \citep{packer2023memgpt}, and \textbf{Hindsight} \citep{latimer2025hindsight}, which index, graph, page, or summarize prior interactions. Beyond these memory architectures, RAG approaches embed documents in a vector store and retrieve relevant chunks at query time. Modern long-context models with 100K to 1M+ token windows place the full context in a single prompt. Each strategy trades accuracy against cost, but the fundamental question remains the same: \textit{does the agent remember not just the facts, but what each of its conclusions depends on?}

  We introduce \textbf{RECON} (\textbf{R}easoning over \textbf{E}xtended \textbf{C}ontexts with \textbf{O}bfuscated \textbf{N}arratives), a benchmark which evaluates whether agents can maintain a coherent, evolving understanding over long contexts where facts don't just accumulate but also interact, contradict, and cascade. RECON spans 24 case files and 1{,}604 questions across three domains (criminal, medical, and financial), with each case ranging from 50k to 100k tokens and fully deterministic ground truth. It tests agents on six memory-intensive tasks, illustrated in Figure~\ref{fig:overview}.

  \begin{enumerate}[noitemsep, topsep=0pt]
      \item \textbf{Chain Reconstruction}: locate and causally order 5--15 evidence hops scattered across the document.
      \item \textbf{Cascade Propagation}: after an invalidation event, determine which conclusions break and which survive via independent support.
      \item \textbf{Source Conflict Resolution}: adjudicate contradictory accounts using independent corroborating evidence.
      \item \textbf{Counterfactual Reasoning}: determine what would have changed under an alternative timeline.
      \item \textbf{Temporal Constraint Satisfaction}: cross-reference parallel data streams against a time window.
      \item \textbf{Temporal Fact Retrieval}: baseline temporal ordering, state queries, and elimination (control).
  \end{enumerate}
    Existing memory benchmarks evaluate whether agents can retrieve scattered facts or detect if a fact has changed, modeling memory as a \textit{state machine} in which facts have current values to be tracked. But tracking current values is not enough: in real-world workflows, facts interact, invalidate, and cascade through explicit dependency structures.

  \paragraph{Contributions.}
  \begin{itemize}[noitemsep, topsep=0pt]
      \item \textbf{The RECON benchmark}: 1{,}604 questions across 24 case files (50K--100K tokens) in three domains, with six task types over evolving rather than static evidence. Human validation reports $\kappa=0.69$ on 200 questions.
      \item \textbf{A deterministic generation pipeline}: a six-layer generator builds each case as a provenance graph with explicit invalidations and counterfactuals, with all ground-truth answers derived in code, not by an LLM.
      \item \textbf{An empirical study of long-context LLMs, RAG variants, and memory-augmented agents on RECON}, analyzing performance across task categories and identifying open challenges. We release the dataset, generator, and evaluation harness.
  \end{itemize}

  \section{Related Work}

  Table~\ref{tab:benchmark_comparison} situates RECON among prior long-context and memory benchmarks across the six task types.

  % --- TABLE 1: Moved up here so it renders at the top of Page 3 ---
  \begin{table*}[t]
  \centering
  \footnotesize
  \setlength{\tabcolsep}{4pt}
  \renewcommand{\arraystretch}{0.9}
  \begin{tabular}{llcccccccc}
  \toprule
  \textbf{Benchmark} & \textbf{Domain} & \textbf{Context} & \textbf{Dyn.} & \textbf{CR-c} & \textbf{CP} & \textbf{SC} & \textbf{CF} & \textbf{TC} & \textbf{TFR} \\
  \midrule
  RULER \citep{hsieh2024ruler} & Synthetic & 4K--128K & $\times$ & $\times$ & $\times$ & $\times$ & $\times$ & $\times$ & $\times$ \\
  LongBench v2 \citep{bai2025longbenchv2} & Multi-domain & 8K--2M & $\circ$ & \checkmark & $\times$ & $\times$ & $\times$ & $\times$ & \checkmark \\
  LooGLE v2 \citep{he2025loogle} & Multi-domain & 16K--2M & $\times$ & \checkmark & $\times$ & $\times$ & $\times$ & \checkmark & \checkmark \\
  Loong \citep{wang2024loong} & Fin./Legal/Acad. & 10K--200K+ & $\times$ & \checkmark & $\times$ & $\times$ & $\times$ & $\times$ & \checkmark \\
  LoCoMo \citep{maharana2024locomo} & Personal conv. & $\sim$10K & $\circ$ & \checkmark & $\times$ & $\times$ & $\times$ & \checkmark & \checkmark \\
  LongMemEval \citep{wu2024longmemeval} & Personal conv. & 115K--1.5M & \checkmark & \checkmark & $\times$ & \checkmark & $\times$ & \checkmark & \checkmark \\
  MemAE \citep{hu2025memae} & Conversational & 100K--1.4M & $\circ$ & \checkmark & $\times$ & \checkmark & $\times$ & \checkmark & \checkmark \\
  TRACK \citep{feng2026track} & Wiki/Code/Math & paragraph & \checkmark & \checkmark & $\times$ & \checkmark & $\times$ & $\times$ & $\times$ \\
  MuSR \citep{sprague2024musr} & Synth. narratives & $\sim$1K words & $\times$ & \checkmark & $\times$ & $\times$ & $\times$ & $\times$ & $\times$ \\
  DetectiveQA \citep{xu2025detectiveqa} & Detective novels & 5K--363K & $\times$ & \checkmark & $\times$ & $\times$ & $\times$ & $\times$ & $\times$ \\
  NovelHopQA \citep{gupta2025novelhopqa} & Public novels & 64K--128K & $\times$ & \checkmark & $\times$ & $\times$ & $\times$ & $\times$ & $\times$ \\
  \midrule
  \textbf{RECON (ours)} & \textbf{Multi-domain} & \textbf{50K--100K} & \textbf{\checkmark} & \textbf{\checkmark} & \textbf{\checkmark} & \textbf{\checkmark} & \textbf{\checkmark} & \textbf{\checkmark} & \textbf{\checkmark} \\
  \bottomrule
  \end{tabular}
  \caption{Comparison of RECON with prior benchmarks. \checkmark{}/$\circ$/$\times$ = tested / partial / not tested. \textbf{Dyn.} = documents contain explicit fact-updates within a single case. Task columns: CR-c (Chain Reconstruction), CP (Cascade Propagation), SC (Source Conflict), CF (Counterfactual), TC (Temporal Constraint), TFR (Temporal Fact Retrieval).}
  \label{tab:benchmark_comparison}
  \end{table*}
  % -----------------------------------------------------------------

  \subsection{Long-Context Evaluation}

  Long-context benchmarks evaluate whether models can retrieve and understand information distributed across long inputs. LongBench \citep{bai2024longbench}, RULER \citep{hsieh2024ruler}, and $\infty$-Bench \citep{zhang2024infinity} established early baselines and pushed evaluation beyond 100K tokens; RULER in particular showed that effective context windows fall well short of advertised limits. NoCha \citep{karpinska2024nocha} introduced globally reasoning-intensive true/false claims, on which GPT-4o reaches only 55.8\% pair accuracy. More recent benchmarks (LongBench v2 \citep{bai2025longbenchv2}, HELMET \citep{yen2025helmet}, LooGLE v2 \citep{he2025loogle}) extend this work to multi-hop settings over 100K+ tokens, with LooGLE v2 reporting only 59.2\% on documents up to 2M tokens.

  These benchmarks evaluate comprehension over \textbf{static} documents. RECON is built around documents that evolve over time: as new facts appear and earlier ones are revised or invalidated, agents must track which conclusions still hold.

  \subsection{Memory Benchmarks}

  A second line of work evaluates memory over accumulated histories. LoCoMo \citep{maharana2024locomo} provides 300-turn chat histories with multi-hop, temporal, and adversarial questions. PerLTQA \citep{du2024perltqa} scales to 8{,}593 questions across 3{,}409 dialogues over personal memory. LongMemEval \citep{wu2024longmemeval} targets temporal state tracking and finds that state-of-the-art systems consistently fail at knowledge updates across sessions. MemAE \citep{hu2025memae} frames memory as four competencies (accurate retrieval, test-time learning, long-range understanding, and selective forgetting). TRACK \citep{feng2026track} extends this line by injecting conflicting fact-updates into multi-step reasoning chains, though at paragraph scale and along a single chain rather than over the bipartite dependency structures RECON evaluates.

  Across these benchmarks, memory is modeled as a \emph{state machine} whose facts have current values to be tracked. RECON instead models it as a \emph{provenance graph} in which conclusions have derivation histories: when evidence is invalidated, the agent must determine which conclusions lose their foundation and which remain supported by independent evidence.

  \subsection{Investigative and Narrative Reasoning}

  A third line of work tests multi-hop reasoning over narrative documents, building on the multi-hop QA tradition \citep{yang2018hotpotqa, ho2020wikimultihop, trivedi2022musique}. MuSR \citep{sprague2024musr} introduces $\sim$1{,}000-word synthetic narratives (including murder mysteries) that probe chain-of-thought reasoning, but the documents are short and contain no fact-updates. DetectiveQA \citep{xu2025detectiveqa} and NovelHopQA \citep{gupta2025novelhopqa} use \emph{real} novels of $\sim$100K and 64K--128K tokens with multi-hop questions, but as published texts they contain no controlled invalidation, source conflict, or counterfactual structure. TurnaboutLLM \citep{yuan2025turnaboutllm} draws contradictions from game transcripts; it overlaps with source conflict but is restricted to a single task.

  \section{The RECON Benchmark}

  \subsection{Overview}
  RECON contains 24 cases and 1,604 questions across crime, medical, and financial domains (contexts of 50K--100K tokens). Each case embeds evidence chains, formal invalidations, source conflicts, counterfactual dependencies, and parallel temporal streams, generated deterministically and rendered into natural language. Systems see only the narrated case file.

  \subsection{Generation Pipeline}
  \label{sec:generation}

\begin{figure*}[t]
      \centering
      \includegraphics[width=0.9\textwidth]{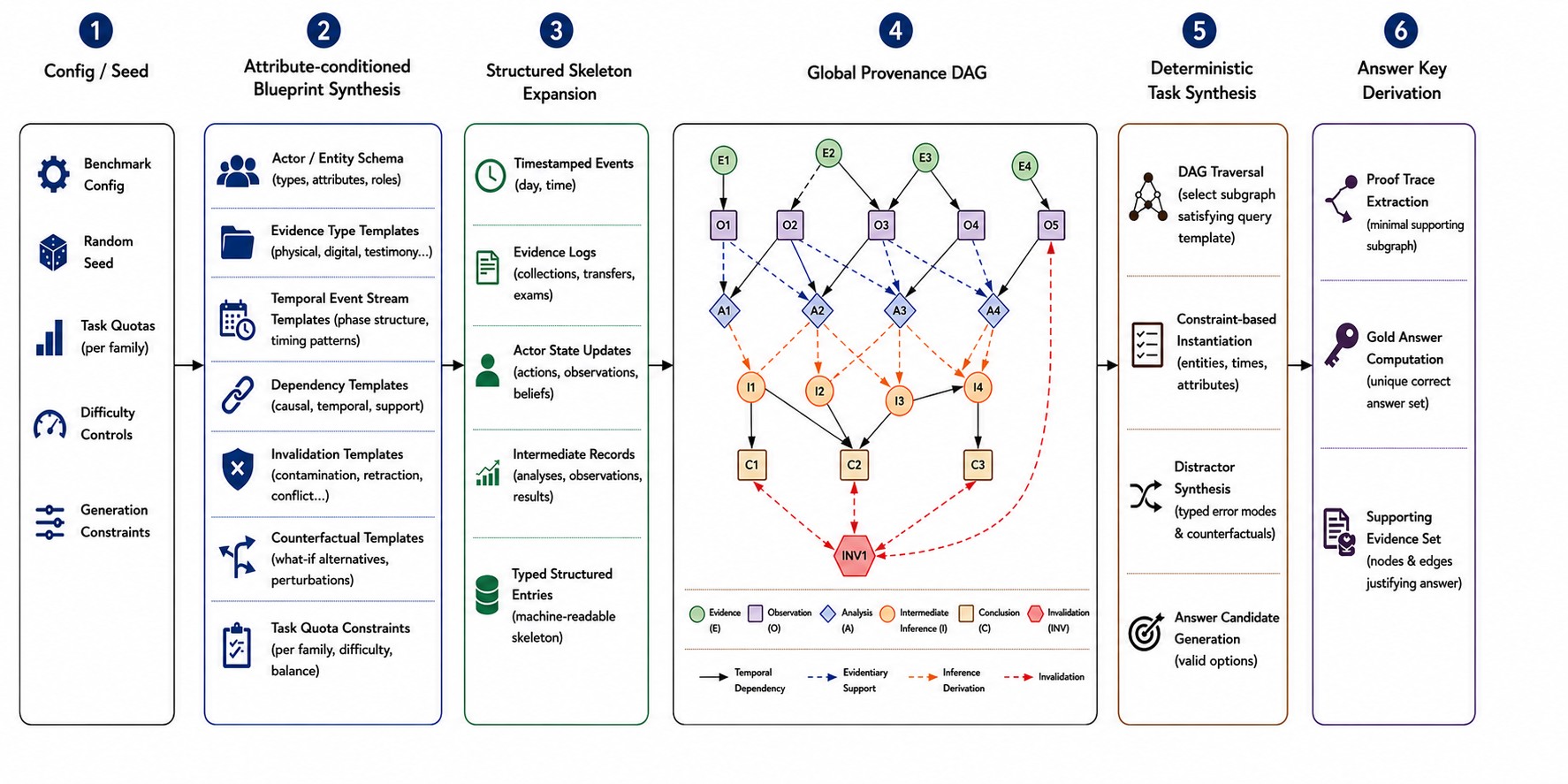}
      \caption{End-to-end RECON generation for a single evidence thread: ground truth is fully established deterministically before LLM narration.}
      \label{fig:worked-pipeline}
  \end{figure*}

Figure~\ref{fig:worked-pipeline} illustrates RECON's deterministic ground-truth construction before narration.

  All case structure is produced by deterministic code. LLMs perform constrained surface realization only and never influence causal structure, provenance, or answer keys; all outputs are postvalidated before release.

  \paragraph{Attribute-conditioned blueprint synthesis.}
  Each case originates from a deterministic, seeded production engine whose
  structure resembles a grammar: a finite catalog of typed terminals (actors,
  locations, evidence templates) is expanded by ordered production rules into a
  complete case blueprint. We formalize the generator as $G = (D, A, P, C)$,
  where $D$ is the domain catalog, $A$ is the set of \emph{attributes} (typed
  fact tags representing the current generation state), $P$ is a set of
  production rules, and $C$ is a set of hard global invariants (temporal
  monotonicity, culprit uniqueness, quota satisfaction). Each rule $p \in P$
  carries a \emph{requires} set $R_p \subseteq A$ and an \emph{establishes} set
  $E_p \subseteq A$: rule $p$ fires only when $R_p \subseteq S_t$ (the current
  state satisfies its preconditions), and on firing updates the state
  $S_{t+1} = S_t \cup E_p$. This contract ensures every generated discovery is
  logically preceded by its prerequisites and induces the provenance DAG directly
  from the generation trace rather than recovering it post-hoc. Given seed $s$, domain $d$, and quota $q$, synthesis is deterministic ($B = G(s,d,q)$).

  \paragraph{Skeleton expansion and temporal enforcement.}
The blueprint expands into a structured skeleton of timestamped events, evidence items, causal dependencies, invalidations, source conflicts, counterfactual dependencies, and parallel temporal streams (surveillance, access, transaction, clinical, or audit). Temporal consistency is enforced by a \textbf{lexicographic monotonic time constraint}: every node carries $\tau(v) = (\text{day}, \text{time})$, and for every causal edge $u \rightarrow v$, $\tau(v) \ge_{\text{lex}} \tau(u)$. A post-skeleton validator verifies these constraints before narration.

  \paragraph{Provenance DAG and proof-trace grounding.}
  The skeleton induces a global provenance DAG whose nodes are events, evidence
  items, and conclusions and whose typed edges encode causal, revisionary,
  invalidating, and conflict-resolving relationships. An edge $u \rightarrow v$
  exists precisely when an attribute established by entry $u$ is required by
  entry $v$, so the graph is read directly off the generation contract rather
  than inferred. This graph is the
  authoritative case representation; narrated prose is downstream of it, not the
  reverse. Every question carries a \textbf{proof trace}: a minimal spanning subgraph of the DAG whose nodes are necessary and sufficient to derive the answer, grounding answer derivation without LLM inference and anchoring the validator's grounding check. Answerability is therefore guaranteed by construction: every gold answer is derivable from this deterministic structure, and a post-question validator enforces uniqueness before release.

  \paragraph{Deterministic task synthesis.}
Questions are generated algorithmically from the provenance DAG: chain reconstruction items sample multi-hop paths; cascade items select an invalidation node and compute via reachability which conclusions collapse and which survive; conflict items locate the independent corroborating node; counterfactual items shift a timestamp and recompute reachability; temporal constraint items query stream intersections; temporal fact retrieval probes single-source chronological facts. A fixed question matrix (Table~\ref{tab:question-distribution}) fixes per-task and per-format quotas. Distractors follow a predefined reasoning-failure taxonomy (Appendix~\ref{sec:appendix_examples}).

  \paragraph{Linguistic realization.}
  After deterministic construction, an LLM narrator converts skeleton entries
  into investigation-style prose under strict fact-fidelity constraints: it may
  choose wording, tone, and sensory detail, but skeleton fields (timestamp,
  source type, entities, evidence identifiers, factual content) are treated as
  immutable. Post-narration validation cross-checks timestamps, entity identifiers, and evidence IDs against the skeleton; failures are re-narrated before release.

  \subsection{Task Categories}
  RECON comprises six memory-intensive task categories. The taxonomy is fixed across domains, while the concrete entities, evidence types, and temporal streams vary by case.

  \paragraph{Chain Reconstruction.}
  The model identifies and causally orders a complete 5--15-hop evidence chain. Adjacent hops are lexically dissimilar and separated by many intervening entries.

  \paragraph{Cascade Propagation.}
  Given a formal invalidation (contaminated sample, retracted statement, corrected audit record, revised clinical finding), the model determines which conclusions lose support and which remain valid through independent evidence. 

  \paragraph{Source Conflict Resolution.}
  The model resolves incompatible accounts by locating independent corroborating evidence. Resolving evidence is unlabeled and often distant in the narrative, so surface heuristics fail.

  \paragraph{Counterfactual Reasoning.}
  The model determines how downstream events change if an earlier discovery occurred at a different time or not at all. The model must keep actual and hypothetical timelines distinct.

  \paragraph{Temporal Constraint Satisfaction.}
  The model satisfies constraints over parallel temporal streams (surveillance, access, transactions, communications, sensor, clinical, audit). 

  \paragraph{Temporal Fact Retrieval.}
  A control category covering lower-composition temporal memory: event ordering, point-in-time state, stream matching, and elimination by explicit evidence. 

  \subsection{Benchmark Validation Pipeline}
  \label{sec:benchmark_validation}

  RECON uses an iterative audit--repair--revalidation pipeline; all validation numbers below refer to the released post-repair benchmark.

  \paragraph{Six-stage validation.}
  Validation has six stages: (1) regex/schema checks for malformed structures and identifier leakage; (2) structural validation (DAG acyclicity, temporal monotonicity, referential integrity); (3) provenance-grounding verification; (4) LLM-assisted semantic audit (diagnostic only, does not define ground truth); (5) robustness audits under anonymization, timestamp normalization, and DAG-safe perturbations; (6) human validation on a stratified sample.

  \paragraph{Released benchmark validation.}
  All 1,604 items pass structural and temporal-consistency validation (0 cyclic dependencies, 100\% temporal monotonicity); average provenance depth is 9 hops (crime) and 6 hops (medical, finance).

  \paragraph{Stress and shortcut validation.}
  Stress tests detected 100\% of injected timestamp inversions, DAG corruptions, and identifier-leakage injections. Shortcut audits confirmed answer stability under anonymization, timestamp normalization, and DAG-safe perturbations, reducing the risk that performance reflects lexical cues rather than provenance reasoning.

  \paragraph{Human validation.}
We validated a stratified sample of 200 questions spanning all domains and task families, with three independent blind annotations per question from a pool of 23 annotators. Annotators were blind to benchmark gold answers and to each other's judgments.

The majority judgment matched the benchmark answer in 86.3\% of cases (Fleiss' $\kappa = 0.69$). Qualitative ratings were positive for 89.7\% of questions on clarity, 91.8\% on evidence sufficiency, and 90.0\% on narrative faithfulness; 13.2\% of items were flagged as ambiguous, concentrated in counterfactual and cascade tasks where upstream invalidations alter downstream conclusions. Full protocol appears in Appendix~\ref{sec:appendix_validation}.

  \section{Evaluation and Results}
  \label{sec:evaluation}

    \subsection{Experimental Setup}
    \label{sec:setup}

    \label{sec:systems}We evaluate three system families plus an Oracle ceiling. The long-context family places the full case file ($\sim$100K tokens) in context, evaluated across eight closed-source and open-weight models. The Oracle replaces the narrated case file with the structured ground-truth representation from which it was generated. The RAG family covers four retrieval variants of increasing sophistication, ranging from plain dense retrieval to hybrid retrieval (dense and BM25) combined with reranking and multi-query expansion. The memory family covers Mem0~\citep{chhikara2025mem0}, Mem0-Graph, Supermemory, and Hindsight~\citep{latimer2025hindsight}.

    \textbf{Scoring.} Every question format uses an explicit scoring rule with an abstain option presented in the prompt. MCQ-Single items use a $-0.2$ penalty for wrong commits, MCQ-Multiple uses the per-option weighted score $c/k - w/(n{-}k)$, Integer items use exact match, Ordering items are scored by Kendall's $\tau$, and free-form responses are graded by two independent LLM judges from different model families (\texttt{gpt-4o} and \texttt{gemini-2.5-flash}); the free-form score is the mean of their two binary verdicts. Exact formulas appear in Appendix~\ref{sec:appendix_eval}.

    \textbf{Metrics.} Our primary metric is Score, the mean of the per-question scoring rule across all evaluated questions. We also report Accuracy, defined as strict correctness with no partial credit or negative marking, which separates raw correctness from calibration. Abstain\% is the fraction of questions for which the system returned no answer, and Tokens/Q is the mean prompt and completion tokens per question.

    \textbf{Contamination filter.} Each of the 1{,}604 questions is first attempted closed-book by three LLMs (\texttt{gpt-5.1}, \texttt{gpt-4.1-mini}, and \texttt{gemini-2.5-flash}), and a question is flagged as priors-guessable when at least two of the three answer it correctly. This procedure flagged 190 questions (11.8\%), leaving 1{,}414 questions for the main evaluation.

    \textbf{Reporting.} Long-context and Oracle results are reported per answering LLM; RAG and memory rows report the unweighted mean across \texttt{gpt-5.1} and \texttt{gemini-2.5-flash} (per-LLM breakdowns and configurations in Appx~\ref{sec:appendix_eval}).

  % -----------------------------------------------------------------------------
  % TABLE 3 (BIG): System x 6 task categories + Overall metrics.
  % RAG / Memory rows averaged across gpt-5.1 + gemini-2.5-flash.
  % -----------------------------------------------------------------------------
  \begin{table*}[t!]
  \centering
  \footnotesize
  \setlength{\tabcolsep}{3pt}
  \renewcommand{\arraystretch}{0.88}
  \resizebox{\textwidth}{!}{%
  \begin{tabular}{llcccccc|cccc}
  \toprule
  & & \multicolumn{6}{c|}{\textbf{Per-Task Score}} & \multicolumn{4}{c}{\textbf{Overall}} \\
  \cmidrule(lr){3-8} \cmidrule(lr){9-12}
  Family & System & Casc. & Chain & CF & SrcConf & TConstr & TFact & Score & Acc.\% & Abst.\% & Tok./Q \\
  \midrule
  \multirow{8}{*}{Long-context}
  & \texttt{gpt-5.1}          & 0.297 & 0.249 & 0.210 & 0.319 & 0.224 & 0.427 & \textbf{0.287} & 20.3\% & 13.6\% & 64{,}649 \\
  & \texttt{gpt-4o-mini}      & \textbf{0.361} & \textbf{0.295} & 0.190 & \textbf{0.343} & 0.043 & 0.161 & 0.232 & 11.2\% &  4.7\% & 64{,}632 \\
  & \texttt{gpt-4.1-mini}     & 0.330 & 0.284 & 0.227 & 0.241 & 0.019 & 0.129 & 0.206 & 10.7\% &  3.8\% & 64{,}635 \\
  & \texttt{gemini-2.5-pro}   & 0.224 & 0.188 & 0.273 & 0.267 & 0.185 & 0.454 & 0.265 & \textbf{22.4\%} & 37.6\% & 70{,}097 \\
  & \texttt{gemini-2.5-flash} & 0.285 & 0.117 & \textbf{0.305} & 0.156 & \textbf{0.237} & \textbf{0.490} & 0.256 & 15.3\% & 11.7\% & 68{,}022 \\
  & \texttt{kimi-k2.5}        & 0.300 & 0.201 & 0.212 & 0.248 & 0.148 & 0.455 & 0.257 & 13.8\% &  8.7\% & 65{,}212 \\
  & \texttt{qwen3-vl-30b}     & 0.315 & 0.232 & 0.232 & 0.269 & 0.142 & 0.208 & 0.230 & 14.6\% & 12.8\% & 70{,}064 \\
  & \texttt{llama-3.3-70b}    & 0.330 & 0.214 & 0.140 & 0.224 & 0.077 & 0.192 & 0.192 &  9.8\% & 11.4\% & 66{,}456 \\
  \midrule
  \multirow{2}{*}{Oracle}
  & \texttt{gpt-5.1}          & 0.923 & 0.792 & 0.474 & 0.935 & \textbf{0.384} & 0.329 & 0.638 & 53.7\% &  2.8\% & 78{,}561 \\
  & \texttt{gemini-2.5-flash} & \textbf{0.939} & \textbf{0.875} & \textbf{0.483} & \textbf{0.961} & 0.290 & \textbf{0.348} & \textbf{0.654} & \textbf{54.6\%} &  6.8\% & 94{,}648 \\
  \midrule
  \multirow{4}{*}{RAG}
  & plain & 0.285 & 0.081 & 0.196 & 0.048 & \textbf{0.177} & 0.353 & 0.178 & 10.5\% & 48.6\% & 7{,}538 \\
  & $+$rerank               & 0.276 & 0.073 & \textbf{0.198} & \textbf{0.119} & 0.150 & 0.403 & \textbf{0.192} & \textbf{14.5\%} & 47.7\% & 7{,}540 \\
  & hybrid $+$ rerank       & \textbf{0.308} & 0.075 & 0.157 & 0.096 & 0.064 & \textbf{0.443} & 0.179 & 11.8\% & 44.5\% & 9{,}037 \\
  & hybrid $+$ rerank $+$ MQ & 0.261 & \textbf{0.101} & 0.174 & 0.094 & 0.143 & 0.315 & 0.173 & 11.5\% & 46.5\% & 8{,}253 \\
  \midrule
  \multirow{4}{*}{Memory}
  & Mem0          & 0.525 & 0.091 & $-0.013$ & 0.005 & 0.124 & 0.101 & 0.112 &  9.7\% & 41.2\% & 3{,}249 \\
  & Mem0-Graph    & 0.572 & 0.048 & 0.071 & 0.096 & \textbf{0.158} & 0.131 & 0.149 & 10.8\% & 45.3\% & 4{,}056 \\
  & Supermemory   & \textbf{0.708} & 0.162 & \textbf{0.207} & \textbf{0.154} & 0.144 & \textbf{0.181} & \textbf{0.211} & \textbf{15.1\%} & 47.2\% & 6{,}440 \\
  & Hindsight     & 0.490 & \textbf{0.197} & $-0.007$ & 0.012 & 0.136 & 0.175 & 0.148 & 10.2\% & 45.6\% & 8{,}438 \\
  \bottomrule
  \end{tabular}%
  }
  \caption{Main results on 1{,}414 clean questions.}
  \label{tab:main-results}
  \end{table*}

  % -----------------------------------------------------------------------------
  % TABLE 4: Oracle--LLM Accuracy gap by task. (Per-domain Score moved to
  % Appendix B, Table~\ref{tab:by-domain}.)
  % -----------------------------------------------------------------------------
  \begin{table}[t!]
  \centering
  \footnotesize
  \setlength{\tabcolsep}{4pt}
  \renewcommand{\arraystretch}{0.95}
  \begin{tabular}{lccc}
  \toprule
  Task & Oracle (best) & Best LLM & Gap (pp) \\
  \midrule
  Source Conflict & 88.9\% & 27.8\% & \textbf{$+$61.1} \\
  Cascade Prop.   & 81.8\% & 18.2\% & \textbf{$+$63.6} \\
  Chain Recon.    & 77.3\% & 18.2\% & \textbf{$+$59.1} \\
  Counterfactual  & 41.2\% & 23.5\% & $+$17.6 \\
  Temp. Constraint & 33.3\% & 20.0\% & $+$13.3 \\
  Temp. Fact Retr. & 23.5\% & 35.3\% & $-$11.8 \\
  \bottomrule
  \end{tabular}
  \caption{Oracle--LLM Accuracy gap by task category. Per-domain Score per system in Table~\ref{tab:by-domain}.}
  \label{tab:oracle-gap}
  \end{table}

  \subsection{Main Results}
  \label{sec:main-results}

  RECON proves difficult for every architecture we evaluate. No non-Oracle exceeds 25\% Accuracy, and even the Oracle, which receives the structured dependency graph in place of the narrated case file, answers fewer than $55\%$ of questions correctly. The best non-Oracle Score is $0.287$ (GPT-5.1), the best non-Oracle Accuracy is $22.4\%$ (Gemini-2.5-Pro), and Oracle reaches $0.654$ and $54.6\%$.

  \textbf{Memory systems excel on Cascade Propagation.} Supermemory's Cascade Score of $0.708$ nearly doubles the best long-context score (GPT-4o-mini $0.361$) and exceeds every RAG variant by roughly $0.4$, while the other three memory systems also reach $0.49$--$0.57$ on this task.

  \textbf{Counterfactual reasoning is the hard ceiling.} The best non-Oracle Counterfactual Score is $0.305$ (Gemini-2.5-Flash), and even the Oracle reaches only $0.483$ despite receiving the dependency graph that defines the case. The bottleneck on this task is therefore chained inference rather than retrieval. Temporal Constraint exhibits the same pattern at lower absolute levels (Oracle $0.384$, best LLM $0.237$).

  \textbf{Retrieval is 8--20$\times$ more token-efficient than long context.} Long-context LLMs consume $64$K--$70$K tokens per question, RAG variants $7$K--$9$K, and memory systems $3$K--$8$K. Despite this reduction, memory matches or exceeds long-context on Cascade (0.49--0.71 vs.\ 0.22--0.36), and RAG comes within 5 pp on Temporal Fact Retrieval (best RAG $0.443$ vs.\ best LC $0.490$). Compression hurts Chain Reconstruction and Source Conflict by discarding inter-fact edges.

  \textbf{Different architectures win different tasks.} Reading off the family-mean Score per task (Figure~\ref{fig:per-task-bars}), long-context leads on the four multi-source reasoning tasks (Chain Reconstruction, Source Conflict, Counterfactual, and Temporal Constraint), RAG leads on Temporal Fact Retrieval (RAG mean $0.378$ vs.\ long-context $0.314$; best variant hybrid$+$rerank reaches $0.443$), and memory leads on Cascade Propagation (Supermemory, $0.708$). No representation handles all six tasks.

  \subsection{Retrieval versus reasoning}
  \label{sec:oracle-gap}

  We decompose system failures along two complementary diagnostic axes: the first compares per-task Accuracy against an Oracle that receives the ground-truth dependency graph as a retrieval-perfect upper bound; the second conditions Accuracy on whether each individual question received its complete gold evidence in the retrieved context.

  The Oracle--LLM Accuracy gap in Table~\ref{tab:oracle-gap} reveals three regimes. Source Conflict, Cascade Propagation, and Chain Reconstruction show large gaps ($+59$ to $+64$ pp), indicating that retrieval is the dominant bottleneck. Counterfactual and Temporal Constraint show much smaller gaps ($+13$ to $+18$ pp) because Oracle itself reaches only $41.2\%$ and $33.3\%$: reasoning limits performance even with perfect context. Temporal Fact Retrieval produces a \emph{negative} $-12$ pp gap because the structured representation abstracts away surface narrative detail this task probes.

  \begin{figure}[t]
  \centering
  % Figure 6: Retrieval-recall decomposition.
% Each family's bar shows the fraction of questions in three exclusive cells:
%   green  = retrieved evidence AND answer correct
%   orange = retrieved evidence AND answer wrong/abstain (reasoning failure)
%   gray   = retrieval miss (retrieval failure)
% Right-side annotations report conditional Accuracy: hit / miss.
%
% Data source: eval_dem/reports/section_4_6_recall.md (overall by family,
% entry-ID recall, EM/full-coverage hit rule). Regenerate via
% eval_dem/scripts/section_4_6_recall.py (writes figures/_recall_data.txt).
% Hit = full coverage of a question's gold supporting entries, anchored on the
% composite [Day X, HH:MM] [TYPE] [AUTHOR] header (near-unique per entry).
% Oracle and Long-context have trivial 100% recall by construction.
% RAG family averages 4 retrieval variants x 2 answering LLMs.
% Memory row is Supermemory (gemini-2.5-flash); Mem0/Mem0-Graph excluded
% because their atomic-fact representations drop the composite headers.

\definecolor{segCorrect}{RGB}{27,158,119}    % teal-green
\definecolor{segWrong}{RGB}{217,95,2}         % orange (reasoning failure)
\definecolor{segMiss}{RGB}{120,120,120}       % gray (retrieval failure)

\begin{tikzpicture}
\begin{axis}[
    xbar stacked,
    width=0.85\columnwidth,
    height=4.5cm,
    bar width=11pt,
    xmin=0, xmax=100,
    xtick={0,25,50,75,100},
    xlabel={Fraction of questions (\%)},
    xlabel style={font=\small, yshift=2pt},
    xticklabel={\pgfmathprintnumber{\tick}\%},
    ytick=data,
    symbolic y coords={Memory,RAG,Long-ctx,Oracle},
    yticklabel style={font=\small},
    xticklabel style={font=\footnotesize},
    ymajorgrids=false,
    xmajorgrids=true,
    major grid style={dashed,gray!25,line width=0.3pt},
    axis line style={line width=0.5pt},
    tick style={line width=0.4pt,black!60},
    enlarge y limits=0.25,
]
% Retrieved AND correct (green) — left segment
\addplot+[xbar, fill=segCorrect, draw=segCorrect!80!black,
    point meta=explicit symbolic,
    nodes near coords={\pgfplotspointmeta},
    nodes near coords style={font=\tiny, color=white, inner sep=1pt, anchor=center}]
    coordinates {(54.2,Oracle) [54] (14.8,Long-ctx) [15] (11.4,RAG) [11] (8.5,Memory) [9]};
% Retrieved AND wrong/abstain (orange) — middle segment
\addplot+[xbar, fill=segWrong, draw=segWrong!80!black,
    point meta=explicit symbolic,
    nodes near coords={\pgfplotspointmeta},
    nodes near coords style={font=\tiny, color=white, inner sep=1pt, anchor=center}]
    coordinates {(45.8,Oracle) [46] (85.2,Long-ctx) [85] (44.0,RAG) [44] (42.6,Memory) [43]};
% Retrieval miss (gray) — right segment (empty label suppresses display for 0% segments)
\addplot+[xbar, fill=segMiss, draw=segMiss!70!black,
    point meta=explicit symbolic,
    nodes near coords={\pgfplotspointmeta},
    nodes near coords style={font=\tiny, color=white, inner sep=1pt, anchor=center}]
    coordinates {(0.0,Oracle) [] (0.0,Long-ctx) [] (44.7,RAG) [45] (48.9,Memory) [49]};
\end{axis}
\end{tikzpicture}
  \caption{Per-family decomposition of questions into retrieval-and-reasoning outcome cells, using full-coverage entry-ID recall. \textcolor[RGB]{27,158,119}{\textbf{Green}}~$=$~retrieval hit and answer correct; \textcolor[RGB]{217,95,2}{\textbf{orange}}~$=$~retrieval hit but wrong/abstain; \textcolor[RGB]{120,120,120}{\textbf{gray}}~$=$~retrieval miss.}
  \label{fig:recall-decomposition}
  \end{figure}
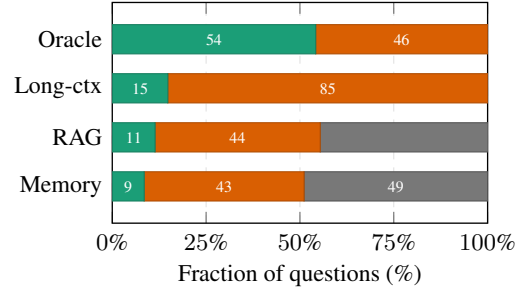

  Following the supporting-fact recall standard in multi-hop QA \citep{yang2018hotpotqa, trivedi2022musique}, we measure for each question the fraction of its gold supporting entries whose composite header \texttt{[Day~X,~HH:MM]~[TYPE]~[AUTHOR]} appears verbatim in the retrieved context. The composite header raises gold-entry anchor uniqueness from $80\%$ to $93\%$ overall ($98\%$ for the multi-hop families). Because supporting evidence is conjunctive for multi-hop tasks, we count a question as a retrieval hit only under full coverage.\footnote{We exclude Mem0 and Mem0-Graph, whose ingestion paraphrases evidence into atomic facts that drop these headers; their overall Score remains in Table~\ref{tab:main-results}.} RAG attains $20.6\%$ Accuracy on full-coverage hits ($n{=}782$) and $1.3\%$ on misses ($n{=}632$), a $19.3$~pp gap. Memory shows the same direction ($16.7\%$ vs.\ $4.3\%$, $n{=}722/692$). Retrieval is necessary but insufficient: roughly four in five full-coverage questions are still answered incorrectly, and on Source Conflict RAG abstains on $47.7\%$ despite holding the conflicting evidence. Long-context holds every entry yet reaches only $14.8\%$ Accuracy (family mean) while Oracle reaches $54.2\%$, a $39.4$~pp gap reflecting the reasoning improvement provided by the structured representation. Reasoning emerges as the dominant bottleneck once retrieval is controlled for.

  \subsection{Human baseline}
  \label{sec:human-baseline}

  We collect a human performance baseline on a 60-question stratified sample (10 per task category, balanced across domains). Three annotators per question received a question-specific evidence packet (the answer-relevant provenance subgraph and matching narrative excerpts) but no oracle aids; we report the majority vote. Annotators reach $63.0\%$ Accuracy (Score $0.54$, abstention rate $11.8\%$, mean $8.8$~minutes per question), exceeding the Oracle solver by $8.4$~pp and pointing to reasoning rather than evidence selection as the residual bottleneck. Per-task and per-domain breakdowns and the full annotation protocol are in Appendix~\ref{sec:appendix_human}.

  \subsection{Testing for narrator-family bias}
  \label{sec:narrator-fingerprint}

  Because our generator rotates four LLM families as narrators across the per-domain seeds (\S\ref{sec:generation}), a solver could in principle be advantaged on cases narrated by its own family. For each LLM that serves as both a narrator and a solver, we measure the Accuracy difference between same-family and cross-family cases using a paired permutation test ($10^4$ case-level label shuffles). The observed differences (GPT-5.1 $+1.3$ pp, $p{=}0.89$; Gemini-2.5-Pro $+13.3$ pp, $p{=}0.32$; Llama-3.3-70B $-5.7$ pp, $p{=}0.47$) are statistically non-significant and inconsistent in direction, indicating no systematic narrator-family advantage.

  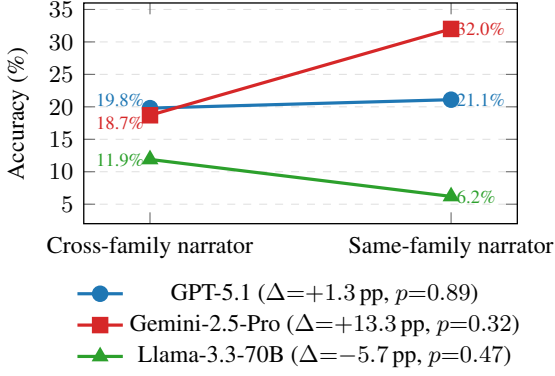
\begin{figure}[t]
  \centering
  % Narrator-fingerprint slope chart.
% For each LLM that doubles as a narrator and a solver, accuracy on the
% cases it narrated (own) vs. accuracy on the rest (other), with
% paired-permutation p-values (10K case-level label shuffles).
%
% Numbers from eval_dem/reports/section_4_4_narrator_fingerprint.md.

\definecolor{lineGPT}{RGB}{31,119,180}   % blue
\definecolor{lineGem}{RGB}{214,39,40}    % red
\definecolor{lineLla}{RGB}{44,160,44}    % green

\begin{tikzpicture}[trim axis left, trim axis right]
\begin{axis}[
    width=0.95\columnwidth,
    height=4.5cm,
    xmin=-0.22, xmax=1.22,
    ymin=2, ymax=36,
    xtick={0,1},
    xticklabels={Cross-family narrator, Same-family narrator},
    xticklabel style={font=\small, yshift=-1pt},
    ytick={5,10,15,20,25,30,35},
    ylabel={Accuracy (\%)},
    ylabel style={font=\small},
    yticklabel style={font=\small},
    ymajorgrids=true,
    major grid style={dashed,gray!25,line width=0.3pt},
    axis line style={line width=0.5pt},
    tick style={line width=0.4pt,black!60},
    legend style={
        font=\footnotesize,
        at={(0.5,-0.22)},
        anchor=north,
        legend columns=1,
        draw=none,
        fill=none,
        /tikz/every even column/.append style={column sep=0.4cm},
    },
    every axis plot/.append style={line width=1.2pt, mark size=2.5pt, mark options={solid}},
]
\addplot[color=lineGPT, mark=*, mark options={fill=lineGPT}]
    coordinates {(0,19.8) (1,21.1)};
\addlegendentry{GPT-5.1 ($\Delta{=}{+}1.3$\,pp, $p{=}0.89$)}

\addplot[color=lineGem, mark=square*, mark options={fill=lineGem}]
    coordinates {(0,18.7) (1,32.0)};
\addlegendentry{Gemini-2.5-Pro ($\Delta{=}{+}13.3$\,pp, $p{=}0.32$)}

\addplot[color=lineLla, mark=triangle*, mark options={fill=lineLla}]
    coordinates {(0,11.9) (1,6.2)};
\addlegendentry{Llama-3.3-70B ($\Delta{=}{-}5.7$\,pp, $p{=}0.47$)}

% Endpoint value labels.
% Left side: GPT (19.8) and Gemini (18.7) are close --- nudge to separate.
\node[font=\scriptsize, color=lineGPT, anchor=east, yshift=3pt, inner sep=2pt]
    at (axis cs:0,19.8) {19.8\%};
\node[font=\scriptsize, color=lineGem, anchor=east, yshift=-3pt, inner sep=2pt]
    at (axis cs:0,18.7) {18.7\%};
\node[font=\scriptsize, color=lineLla, anchor=east, inner sep=2pt]
    at (axis cs:0,11.9) {11.9\%};

\node[font=\scriptsize, color=lineGPT, anchor=west, inner sep=2pt]
    at (axis cs:1,21.1) {21.1\%};
\node[font=\scriptsize, color=lineGem, anchor=west, inner sep=2pt]
    at (axis cs:1,32.0) {32.0\%};
\node[font=\scriptsize, color=lineLla, anchor=west, inner sep=2pt]
    at (axis cs:1,6.2) {6.2\%};
\end{axis}
\end{tikzpicture}
  \caption{Same-family vs.\ cross-family narrator Accuracy for each LLM that doubles as narrator and solver.}
  \label{fig:narrator-fingerprint}
  \end{figure}

  \subsection{Performance scales inversely with chain length}
  \label{sec:complexity-gradient}

\begin{figure}[t]
  \centering
  % Figure 5: Accuracy on Chain Reconstruction by chain length, per family.
% Restricted to chain_reconstruction questions.
% Accuracy (not Score) is reported because Score is sensitive to abstain rate.
%
% RAG and Memory x-coordinates are slightly jittered (+0.10 / -0.10) so their
% markers and lines are visually distinguishable when both sit at 0% accuracy.
% Long-context and Oracle stay at integer x.
%
% Data source: paper/figures/_complexity_data.txt (generated by
% eval_dem/scripts/section_4_7_complexity.py).

\definecolor{lineOracle}{RGB}{35,75,115}     % deep navy
\definecolor{lineLong}{RGB}{217,95,2}         % orange
\definecolor{lineRAG}{RGB}{27,158,119}        % teal
\definecolor{lineMem}{RGB}{117,112,179}       % purple

\begin{tikzpicture}
\begin{axis}[
    width=0.85\columnwidth,
    height=5.0cm,
    xmin=2.5, xmax=10.5,
    ymin=-3, ymax=110,
    xtick={3,4,5,6,7,10},
    xticklabels={3,4,5,6,7,10},
    ytick={0,25,50,75,100},
    xlabel={Chain length (\# of evidence hops required)},
    ylabel={Accuracy (\%)},
    xlabel style={font=\small},
    ylabel style={font=\small},
    xticklabel style={font=\small},
    yticklabel style={font=\small},
    ymajorgrids=true,
    major grid style={dashed,gray!25,line width=0.3pt},
    axis line style={line width=0.5pt},
    tick style={line width=0.4pt,black!60},
    legend style={
        font=\footnotesize,
        at={(0.5,1.03)},
        anchor=south,
        legend columns=2,
        column sep=0.4cm,
        row sep=0.05cm,
        draw=none,
        fill=none,
    },
    every axis plot/.append style={
        line width=1.1pt,
        mark size=2.5pt,
        mark options={solid},
    },
]
% Oracle (no x-jitter)
\addplot[color=lineOracle, mark=*, mark options={fill=lineOracle}]
    coordinates {
        (3,  100.0) (4,  100.0) (5,  75.0)
        (6,  62.5)  (7,  78.6)  (10, 50.0)
    };
\addlegendentry{Oracle}

% Long-context (no x-jitter)
\addplot[color=lineLong, mark=square*, mark options={fill=lineLong}]
    coordinates {
        (3,  0.0)  (4,  0.0)  (5,  12.5)
        (6,  3.1)  (7,  10.7) (10, 15.6)
    };
\addlegendentry{Long-context}

% RAG (x-jitter +0.10 to separate from Memory at 0% lines)
\addplot[color=lineRAG, mark=triangle*, mark options={fill=lineRAG}]
    coordinates {
        (3.10,  0.0) (4.10,  0.0) (5.10,  0.0)
        (6.10,  0.0) (7.10,  0.0) (10.10, 9.4)
    };
\addlegendentry{RAG}

% Memory (x-jitter -0.10 to separate from RAG at 0% lines)
\addplot[color=lineMem, mark=diamond*, mark options={fill=lineMem}]
    coordinates {
        (2.90, 0.0) (3.90, 0.0) (4.90, 0.0)
        (5.90, 0.0) (6.90, 0.0) (9.90, 8.3)
    };
\addlegendentry{Memory}
\end{axis}
\end{tikzpicture}
  \caption{Accuracy on Chain Reconstruction by chain length, per family.}
  \label{fig:complexity-gradient}
  \end{figure}
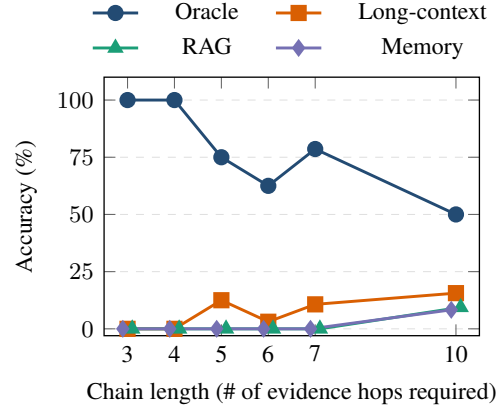

  We restrict the gradient analysis to Chain Reconstruction, whose questions carry an explicit per-question complexity scalar (\texttt{chain\_length}, range 3--10), isolating a within-task signal. Figure~\ref{fig:complexity-gradient} reports Accuracy by chain length per family.

  Oracle Accuracy declines monotonically, from $100\%$ at chain length 3--4 to $50\%$ at chain length 10: even given the full dependency graph, step-by-step composition degrades with depth. Non-Oracle systems remain near zero across all chain lengths (long-context $0$--$16\%$, RAG and memory $0$--$9\%$), failing at the retrieval or attention layer before chain length becomes the operative bottleneck. The widening gap between Oracle and the remaining families confirms that the Oracle ceiling identified in \S\ref{sec:oracle-gap} is itself a function of compositional depth.

  \section{Conclusion}

  We introduced RECON, a benchmark that evaluates whether agents can maintain a coherent, evolving understanding of long documents in which facts interact, contradict, and cascade. RECON departs from prior memory benchmarks by modeling memory as a provenance graph rather than a state machine: conclusions carry derivation histories and invalidation events propagate through explicit dependencies. The benchmark contains $1{,}604$ questions over $24$ case files ($50$K--$100$K tokens) across three domains, constructed by a deterministic generator that places no LLM in the gold reasoning trace. Across long-context, retrieval, and agentic memory architectures, both retrieval and reasoning remain open: an Oracle with the ground-truth dependency graph reaches only $54.6\%$ Accuracy, the best non-Oracle system just $22.4\%$, and roughly four in five failures persist even when retrieval succeeds. We release the benchmark, evaluation harness, and generation code to support future work.

  \section{Limitations}
  \textbf{Synthetic data.} Case narratives are LLM-generated over deterministic ground truth. While narration is constrained to skeleton facts, models trained on similar synthetic narratives may have a stylistic advantage. We mitigate this by generating unique characters and evidence structures, but sensitivity to narrative style cannot be fully ruled out.

  \textbf{Schema leakage.} Knowledge of the internal skeleton schema could allow structured extraction pipelines to exploit the answer format rather than reasoning from the narrative.

  \textbf{Human baseline input asymmetry.} The human baseline (\S\ref{sec:human-baseline}) provides annotators with an answer-relevant evidence packet, whereas LLM solvers operate over the full 100K-token case file or a self-retrieved context. The reported human Accuracy should therefore be interpreted as a complementary upper bound to the Oracle solver, which similarly receives curated evidence in structured form, rather than as a direct comparison against long-context systems.

  \bibliography{custom}

  \onecolumn
  \clearpage
  \appendix

  \section{LLM Prompt Templates Used in RECON}
  \label{sec:appendix_prompts}

  The LLM is confined to surface realization (narration, distractor editing, question rewriting) and to evaluation (answering and judging); it never produces case structure, provenance, or answer keys (\S\ref{sec:generation}). All generation calls run at \texttt{temperature}~$=0$ and decode under a strict JSON schema so that no text can appear outside the intended field. In the templates below, field names are shown in \texttt{typewriter} and braces denote interpolated values.

  \subsection{Narrative Generation Prompt}

  \begin{promptbox}[System Prompt]
  You are a professional writer producing one entry for a \{domain\} case file. Convert the structured FACTS provided into realistic prose, in the register of a \{register\}, written in the past tense. Follow these rules:
  \begin{itemize}[noitemsep,leftmargin=*]
  \item Write only the narrative body of the entry. Do not include planning, drafts, headers, or metadata labels such as ENTRY TYPE, AUTHOR, or DATE; the surrounding case-file template adds these automatically.
  \item Use only the information in the FACTS block. Do not introduce specific details that are not given, including the names of venues or businesses, exact times, monetary amounts, identifiers, quoted speech, telephone numbers, or addresses. Where a specific detail is not provided, describe it generically (for example, ``the claimed location'' or ``earlier that evening'').
  \item Whenever the FACTS block names a person, refer to that person by their full name. Do not anonymize a named individual as ``an associated party'' or ``the subject''. If the FACTS block itself uses a generic noun, retain it.
  \end{itemize}
  The \{domain\}/\{register\} pair is one of: criminal investigation / police investigation report; medical adverse-event investigation / clinical incident report (objective, passive voice, describing systemic events rather than assigning individual blame); financial fraud investigation / regulatory-enforcement or forensic-accounting narrative (precise about dates, amounts, and filing references).
  \end{promptbox}

  \begin{promptbox}[User Prompt]
  \begin{verbatim}
  ENTRY TYPE: {entry type}
  AUTHOR / SOURCE: {author}
  DATE: Day {day}, {time}
  REGISTER: {tone}

  STYLE GUIDE FOR THIS ENTRY TYPE:
  {per-type style guide}

  FACTS (every item must appear; add nothing;
  do not repeat these field names):
  {structured data as a bullet list}

  Write the entry now, as prose only.
  \end{verbatim}
  \end{promptbox}

  \noindent The style guide is drawn from a fixed per-entry-type library (e.g.\ witness statement, forensic lab report, financial record); \{tone\} is sampled from a seeded register list, with automated entry types fixed to neutral. Output is constrained to a JSON object with a single \texttt{entry} field at \texttt{temperature}~$=0$.

  \subsection{Distractor Enhancement Prompt}

  \begin{promptbox}[System Prompt]
  You are a question designer for the RECON investigative-reasoning benchmark. Given a JSON description of a \{domain\} case and a set of multiple-choice questions, improve the incorrect options (distractors) so that each question can be answered only by consulting the case, not from surface cues. Rules:
  \begin{itemize}[noitemsep,leftmargin=*]
  \item Leave the correct option unchanged.
  \item Replace generic or out-of-domain distractors with plausible but incorrect alternatives drawn from the case and appropriate to the question type (e.g.\ for chain reconstruction, incorrect steps from the same or a similar chain; for source conflict, incorrect resolutions of that specific conflict).
  \item Surface uniformity: all options share the same length range, register, specificity, and structure, so the correct option is not identifiable by form alone.
  \item Stem independence: a distractor must not be refutable from the question stem in isolation.
  \item Case vocabulary: phrase options in the language of the narrated case file; do not use pipeline scaffolding terms such as ``trigger event'' or ``chain step''.
  \item Plausibility: each distractor must be credible to a careful reader who has not seen the case.
  \item Within-batch variation: vary distractors and phrasings across questions that share a template.
  \end{itemize}
  Return JSON only: an \texttt{enhanced} array in which each element gives the question \texttt{id}, the full \texttt{options} list (correct text unchanged), and a \texttt{rationale} per wrong option (label, reason, and one failure mode from the fixed taxonomy).
  \end{promptbox}

  \begin{promptbox}[User Prompt]
  \begin{verbatim}
  CASE CONTEXT:
  {case as JSON: entities, causal chains, findings,
  conflicts, counterfactuals, temporal streams}

  QUESTIONS TO ENHANCE:
  {array of MCQ questions}
  \end{verbatim}
  \end{promptbox}

  \subsection{Question Rewriting Prompt}

  \begin{promptbox}[System Prompt]
  You are a question editor for the RECON benchmark. Rewrite generated questions and options so they read naturally to any reader, without changing their meaning.
  \begin{itemize}[noitemsep,leftmargin=*]
  \item Preserve which option is correct, the number and labels of options, and the facts in the reasoning. Do not change the answer.
  \item Remove internal identifiers (e.g.\ \texttt{log\_001}, \texttt{EV-01}) and pipeline terms, replacing them with natural language (``data stream'' becomes ``investigative record''; ``evidence chain'' becomes ``evidence trail''). Render entry descriptions in plain English (``the witness statement by P regarding S'' becomes ``P's witness statement about S'').
  \item Maintain anti-leakage properties: all options share one surface format, length range, and level of detail; no distractor is refutable from the stem alone; option vocabulary matches the case file; phrasings vary across questions that share a template.
  \item Every option ends with a period, and no two options are identical.
  \end{itemize}
  Return JSON only, giving for each question its \texttt{id} and the rewritten question, options, and reasoning; all other fields are preserved downstream.
  \end{promptbox}

  \begin{promptbox}[User Prompt]
  \begin{verbatim}
  Case context (for humanizing names and terms):
  {case summary}

  INPUT QUESTIONS:
  {array of questions}
  \end{verbatim}
  \end{promptbox}

  \subsection{Answering Prompt (All Solvers)}

  Every solver shares one answering template; systems differ only in the descriptor of the supplied context, which is appended to the system message.

  \begin{promptbox}[System Prompt]
  You are an expert investigation analyst. You have been given \{context descriptor\}. Answer the question based only on this information. Be precise and concise.
  \end{promptbox}

  \noindent The \{context descriptor\} identifies the system:
  \begin{itemize}[noitemsep,leftmargin=*]
  \item \textbf{Long-context}: the complete case file.
  \item \textbf{Oracle}: the full structured ground truth for the case, in JSON; every fact needed to answer is present.
  \item \textbf{RAG / Hybrid RAG / Supermemory}: retrieved excerpts from the case file.
  \item \textbf{Mem0 / Hindsight}: memories retrieved from the case file.
  \item \textbf{Mem0-Graph}: retrieved memories and knowledge-graph relations from the case file.
  \item \textbf{Closed-book}: no case file or context is supplied; rely only on prior knowledge and abstain if the answer cannot be determined with confidence.
  \end{itemize}

  \begin{promptbox}[User Prompt]
  \begin{verbatim}
  {question text}
    A) {option}
    B) {option}  ...

  {format instruction}
  \end{verbatim}
  \end{promptbox}

  \noindent The format instruction is one of:
  \begin{itemize}[noitemsep,leftmargin=*]
  \item \textbf{MCQ (single)}: select the single best answer (A--F) in \texttt{answer}; if undetermined, set \texttt{abstain} true and \texttt{answer} null. Abstaining is the only valid way to signal uncertainty. Scoring: correct $+1$, wrong $-0.2$, abstain $0$.
  \item \textbf{MCQ (multiple)}: select all correct options (A--F) in \texttt{answers}; if undetermined, set \texttt{abstain} true with an empty list. Each correct selection adds and each wrong selection subtracts; select only confident options.
  \item \textbf{Integer}: give an exact integer in \texttt{answer}; if undetermined, set \texttt{abstain} true and do not guess a number. Scoring: correct $+1$, otherwise $0$.
  \item \textbf{Ordering}: give the option letters in order in \texttt{order}; if undetermined, set \texttt{abstain} true with an empty list, not a partial or guessed order. A reversed order receives the largest penalty.
  \item \textbf{Freeform}: answer in 2--3 sentences using only the context, in \texttt{answer}; if undetermined, set \texttt{abstain} true. Scoring: correct $+1$, otherwise $0$.
  \end{itemize}

  \subsection{LLM Judge Prompt}

  Freeform answers are graded by two independent LLM judges from different model families (\texttt{gpt-4o} and \texttt{gemini-2.5-flash}); the per-question free-form score is the mean of their two binary verdicts. All other formats are scored deterministically (Appendix~\ref{sec:appendix_eval}). The prompt template below is used for both judges.

  \begin{promptbox}[System Prompt]
  You are a strict grader for a multi-domain investigation-reasoning benchmark. You receive a question, the expected answer (ground truth derived from a deterministic skeleton), and a student answer. Grade the student answer as CORRECT or INCORRECT.
  \begin{itemize}[noitemsep,leftmargin=*]
  \item It is CORRECT if it captures the key factual content of the expected answer; paraphrasing, minor wording differences, and additional correct detail are acceptable.
  \item It is INCORRECT if it omits a key fact, contradicts the expected answer, contains a factual error, or is empty (including ``I don't know'' or ``cannot determine'').
  \end{itemize}
  Respond with exactly one word: CORRECT or INCORRECT.
  \end{promptbox}

  \begin{promptbox}[User Prompt]
  \begin{verbatim}
  QUESTION: {question text}

  EXPECTED ANSWER: {ground-truth answer}

  STUDENT ANSWER: {model response}
  \end{verbatim}
  \end{promptbox}

  \clearpage
  \section{Generation Pipeline: Full Details}
  \label{sec:appendix_generation}

  \begin{figure*}[!ht]
      \centering
      \includegraphics[width=\textwidth]{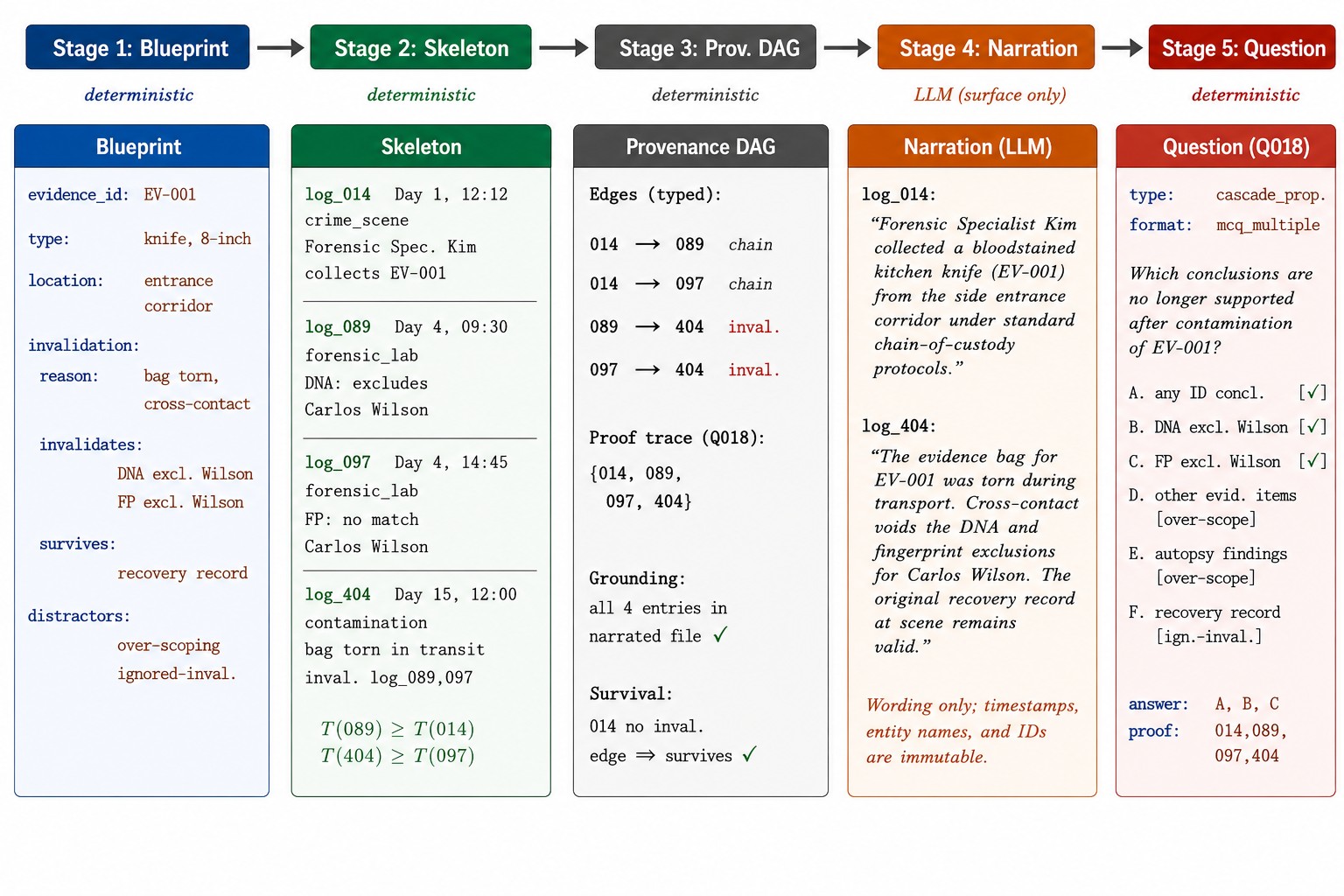}
      \caption{
      Full deterministic RECON generation pipeline. Case construction proceeds
      from seeded configuration through attribute-conditioned blueprint synthesis,
      structured skeleton expansion, provenance DAG induction, deterministic task
      synthesis, and answer derivation. LLMs are restricted to constrained surface
      realization and do not participate in authoritative reasoning or ground
      truth construction.
      }
      \label{fig:appendix-generation-pipeline}
  \end{figure*}

  RECON case files are generated through a deterministic multi-stage pipeline (Figure~\ref{fig:appendix-generation-pipeline}): \textbf{Seeded Configuration} $\to$ \textbf{Attribute-Conditioned Blueprint Synthesis} $\to$ \textbf{Skeleton Expansion} $\to$ \textbf{Provenance DAG Construction} $\to$ \textbf{Task Synthesis} $\to$ \textbf{Narrative Realization} $\to$ \textbf{Validation}. Ground truth is established entirely by deterministic program logic. LLMs are used only for constrained linguistic surface realization and optional lexical polishing, never for defining causal structure, provenance, dependencies, or answer keys.

  \subsection{Stage 1: Attribute-Conditioned Blueprint Construction}

  Each case begins from a deterministic seeded configuration specifying domain, actor cardinalities, evidence quotas, temporal stream types, dependency budgets, invalidation counts, conflict counts, and counterfactual branching constraints.

  An attribute-conditioned production engine, structured analogously to a grammar with typed terminals and ordered rewrite rules, expands this configuration into a typed blueprint schema containing entities, events, evidence classes, dependency templates, and task-relevant structural constraints. Each rule fires only when its required attributes are present in the current generation state and establishes new attributes on firing. Because synthesis is deterministic under a fixed seed, identical inputs always reproduce identical case structures.

  \subsection{Stage 2: Deterministic Skeleton Construction}

  The skeleton and provenance graph jointly define the authoritative ground truth in RECON. They are constructed entirely by deterministic code with no LLM involvement. The skeleton instantiates:

  \begin{itemize}[noitemsep, topsep=2pt]
    \item \textbf{Temporal structure}: a domain-specific temporal timeline with precise timestamps for every event, including time-of-death windows, forensic processing delays, and evidence discovery sequences.
    \item \textbf{Evidence chains}: multi-hop causal chains (5--15 hops) linking initial observations to final investigative conclusions, with each hop assigned to a specific day and log entry type.
    \item \textbf{Invalidation events}: formal contamination or retraction events that sever specific links in the dependency graph. Each invalidation stores the affected evidence node. Downstream impact is not hardcoded as a conclusion list; it is computed dynamically by provenance graph traversal, allowing conclusions with independent support to survive invalidation. This explicit dependency structure is what makes cascade propagation and counterfactual questions answerable by construction: the skeleton knows exactly which conclusions collapse when a given piece of evidence is removed, and which survive through independent support.
    \item \textbf{Parallel data streams}: four independent streams (cell tower pings, surveillance footage, transaction receipts, building access logs) that provide overlapping but incomplete temporal coverage. No single stream covers the full time window, so definitive placement or alibi conclusions require intersection across at least two streams.
    \item \textbf{conflicting source assertions}: pairs of irreconcilable witness accounts, each paired with a designated independent source (from a different data stream) that resolves the conflict.
    \item \textbf{Counterfactual dependencies}: explicit provenance dependencies recording which investigative actions (warrants, suspect links, follow-up interviews) depend on which earlier discoveries. Shifting a discovery's timestamp to an earlier or later point in the timeline deterministically changes which downstream actions would or would not have occurred.
  \end{itemize}

  All question answers are derived directly from these structures. Because the skeleton is constructed deterministically, every answer is correct by construction and the entire benchmark is reproducible from a single seed.

  \subsection{Stage 3: Provenance DAG Construction}

  The expanded skeleton induces a global directed acyclic provenance graph whose nodes represent events, evidence artifacts, assertions, and conclusions. Typed edges encode causal dependence, corroboration, revision, invalidation, and conflict-resolution relationships.

  Question ground truth is derived from graph computation over this structure rather than from narrated text. Cascade propagation is computed by invalidating upstream nodes and recomputing downstream reachability. Source conflict resolution identifies contradictory assertions and independent corroborating support. Counterfactual questions modify intervention variables and recompute affected dependency paths.

  \subsection{Stage 4: Task Synthesis}

  After the provenance DAG is constructed, the question generator algorithmically traverses it to produce questions for each task category (per-task synthesis logic is described in \S\ref{sec:generation}). A fixed question matrix enforces per-task, per-format quotas across all cases, eliminating distributional drift between generation runs. Table~\ref{tab:question-distribution} reports the resulting distribution: 1{,}604 questions across 24 case files in three domains, spanning six task categories and five answer formats.

  \begin{table}[t]
  \centering
  \scriptsize
  \setlength{\tabcolsep}{2pt}
  \renewcommand{\arraystretch}{0.72}
  \begin{tabular}{lcccccc}
  \toprule
  Task & MCQ-S & MCQ-M & Int & Ord & Free & Total \\
  \midrule
  Cascade      & 72 & 72 & 48 & 48  & 48 & 288 \\
  Chain        & 48 & 48 & 24 & 144 & 72 & 336 \\
  Counterfact. & 72 & 72 & 48 & 48  & 24 & 264 \\
  Conflict     & 72 & 72 & 24 & 48  & 48 & 264 \\
  Temp. Const. & 56 & 72 & 48 & 16  & 24 & 216 \\
  Temp. Retr.  & 72 & 48 & 72 & 20  & 24 & 236 \\
  \midrule
  \textbf{Total} & \textbf{392} & \textbf{384} & \textbf{264} & \textbf{324} & \textbf{240} & \textbf{1604} \\
  \bottomrule
  \end{tabular}
  \caption{Question distribution by task and answer format.}
  \label{tab:question-distribution}
  \end{table}

  \subsection{Stage 5: Controlled Narrative Realization}

  An LLM converts immutable structured entries into natural-language case logs under strict fact-preservation constraints. Input fields include timestamp, source type, participating entities, evidence identifiers, and factual payload.

  The narrator may alter wording and discourse structure but may not introduce, remove, or modify factual content. Post-generation validation checks timestamp fidelity, entity preservation, and evidence-reference consistency; failed entries are regenerated.

  The narrator is constrained to the facts established in the skeleton: it may choose phrasing, tone, and level of detail, but it may not introduce new facts, alter timestamps, or change relationships between entities. Narration proceeds entry-by-entry, with each entry receiving the skeleton's structured fields (timestamp, source type, actors involved, factual content) and producing a natural language paragraph. The narrator also receives a sliding window of recently narrated entries to maintain stylistic continuity and avoid repetition.

  This separation is the key design decision in RECON's pipeline. The skeleton guarantees correctness; the narrator guarantees naturalness. A mechanically generated case file would be trivially gameable by pattern matching, while an LLM-generated ground truth would be unreliable. By assigning each concern to the appropriate tool (deterministic code for truth, LLMs for language), RECON achieves both.

  The narration stage produces approximately 800 timestamped log entries spanning 30 days, totaling roughly 100,000 tokens. The resulting case file reads as a coherent investigation narrative, with no visible seams between the deterministic structure underneath and the natural language surface.

  \subsection{Validation}

  The pipeline enforces deterministic structural validation at every stage to ensure correctness before proceeding.

  \textbf{Post-skeleton validation} checks structural integrity: every evidence chain must have the required number of hops with no dangling references, every invalidation event must point to a valid evidence item with at least one dependent conclusion, every witness conflict must have exactly two conflicting accounts and one resolving source, and every counterfactual dependency must form a directed acyclic graph. Temporal constraints are also verified: no event precedes the investigation start, forensic results do not appear before their corresponding evidence is collected, and invalidation events are temporally consistent with the evidence and dependency structures they modify.

  \textbf{Post-narration validation} checks that the narrated case file preserves skeleton facts. Timestamps in the narrated text are extracted and compared against skeleton timestamps. Entity names, evidence item identifiers, and key factual claims are verified to appear in the narrated output. Entries that fail validation are re-narrated with explicit correction instructions.

  \textbf{Post-question validation} checks that every generated question has a unique, unambiguous answer derivable from the skeleton. For MCQ questions, it verifies that exactly one option (or the correct subset) matches the ground truth, that all distractor options are distinct, and that the correct answer label is properly assigned. For ordering questions, it verifies that the expected sequence is a valid topological sort of the underlying dependency graph.

  \clearpage

  \section{Extended Human Validation Analysis}
  \label{sec:appendix_validation}

  \paragraph{Annotation protocol.}
  To assess benchmark quality beyond deterministic structural validation, we conducted human evaluation on a stratified sample of 200 questions covering all three domains (crime, finance, medical) and all task families of the RECON benchmark. Each question received three independent blind annotations from a pool of 23 annotators. Two candidate items exhibiting unresolved structural ambiguity were removed during pre-release validation; reported statistics reflect the released benchmark.

  The annotator pool comprised members of the academic community, including undergraduate and postgraduate students. All participants received standardized written annotation guidelines and completed a calibration phase on held-out examples disjoint from the released benchmark.

  Annotators evaluated five dimensions: benchmark answer correctness, question clarity, evidence sufficiency, narrative faithfulness, and ambiguity. Benchmark answer correctness was evaluated against the released benchmark answer, while qualitative dimensions were recorded as independent binary judgments.

  \paragraph{Agreement and benchmark quality.}
  The majority annotator judgment matched the benchmark answer for 86.3\% of questions. Fleiss' $\kappa = 0.69$, computed over the correctness judgments, indicates substantial inter-annotator agreement. Across qualitative dimensions, 89.7\% of questions were rated clear, 91.8\% were judged to provide sufficient evidence for answer derivation, and 90.0\% were assessed as faithfully aligned with the benchmark's intended narrative and underlying evidence structure. Only 13.2\% of items were flagged as ambiguous.

  \paragraph{Disagreement analysis.}
  Observed disagreements were concentrated in higher-composition reasoning settings, particularly those involving invalidation and counterfactual dependency reasoning. The most common disagreement modes involved:
  \begin{itemize}[noitemsep, topsep=2pt]
      \item failure to integrate later invalidating evidence over earlier plausible evidence,
      \item incorrect preservation of downstream conclusions after upstream invalidation,
      \item confusion between actual and counterfactual timelines,
      \item incomplete multi-hop evidence integration over long contexts.
  \end{itemize}

  For example, in one financial source-conflict case, annotators disagreed on whether a transaction should be treated as legitimate based on an early bank transfer record, whereas the released benchmark answer required incorporating a later fraud alert that invalidated that evidence path. Similar disagreement patterns appeared in counterfactual questions where changing an upstream premise deterministically altered downstream conclusions.

  \paragraph{Relation to deterministic validation.}
  Human validation complements, rather than replaces, RECON's deterministic pre-release validation pipeline, under which every released artifact passes schema, DAG-consistency, temporal-monotonicity, provenance-grounding, and robustness checks; items failing these checks are repaired or regenerated before release. Human disagreement therefore reflects residual reasoning difficulty rather than generation pipeline failure.

  \clearpage
  \section{Evaluation Configurations}\label{sec:appendix_eval}

  This appendix documents the per-architecture configurations referenced in \S\ref{sec:setup}. Long-context and Oracle place the full structured input directly in the prompt; the RAG and memory architectures share a common pre-processing step that splits the case file into approximately $800$ chunks at log-entry boundaries (one chunk per timestamped entry).

  \subsection{Scoring Rules}
  \label{sec:scoring_rules}

  The per-format scoring rules referenced in \S\ref{sec:setup} are defined as follows.

  \begin{itemize}[noitemsep, topsep=2pt, leftmargin=*]
      \item \textbf{MCQ Single} ($n=6$ options): correct $=+1$, wrong $=-1/(n{-}1)=-0.2$, abstain $=0$. With $n=6$, random guessing and abstention both yield an expected score of $0$, while wrong commits incur a strict penalty.
      \item \textbf{MCQ Multiple} ($n=6$ options, $k$ correct): score $= c/k - w/(n{-}k) \in [-1, +1]$, where $c$ and $w$ count correctly- and incorrectly-selected options. Selecting all options yields $0$, identical to abstaining.
      \item \textbf{Integer}: exact match; correct $=+1$, wrong $=0$, abstain $=0$.
      \item \textbf{Ordering}: the solver receives the gold set of $m$ items and outputs a permutation, scored by Kendall's $\tau = (C-D)/\binom{m}{2} \in [-1, +1]$. Outputs containing wrong items or wrong cardinality are assigned $\tau = 0$.
      \item \textbf{Free-form}: two independent LLM judges (\texttt{gpt-4o}, snapshot \texttt{gpt-4o-2024-08-06}, and \texttt{gemini-2.5-flash}) each emit a binary verdict (correct $=1$, incorrect $=0$) over the same response, and the per-question free-form score is the mean of the two: both correct $\to 1$, split $\to 0.5$, both wrong $\to 0$. No negative marking is applied to free-form: unlike MCQ there is no discrete option set against which to calibrate a guessing penalty. An abstention scores $0$.
  \end{itemize}

  \subsection{Determinism and Reproducibility}

  All solver calls run at \texttt{temperature=0}. Every model is snapshot-pinned (e.g., \texttt{gpt-4.1-mini-2025-04-14}, \texttt{gpt-4o-mini-2024-07-18}) so that silent provider-side updates cannot perturb results. Responses are decoded against per-format structured-output schemas, so every output is either a parsed answer or an explicit abstention; parsing failures are retried up to three times and then logged as abstentions. Cloud memory services (Mem0, Mem0-Graph, Supermemory) ingest each case file exactly once per case, and the resulting memory state is reused across all retrievals for that case. This both controls ingestion-side stochasticity and keeps per-question Tokens/Q comparable across systems.

  \subsection{Statistical Reporting and Judge Validation}

  Pairwise comparisons between systems use paired bootstrap over questions with $10^4$ resamples, and we report 95\% confidence intervals on the Score difference. Free-form responses are graded by two independent LLM judges from different model families: \texttt{gpt-4o} (snapshot \texttt{gpt-4o-2024-08-06}), following the LongMemEval ICLR 2025 precedent, and \texttt{gemini-2.5-flash}. Using two cross-family judges is a control against the documented self-preference bias of single-family judges toward their own family's outputs. Each judge independently scores every free-form response, and the per-question free-form score is the mean of the two binary verdicts (Appendix~\ref{sec:scoring_rules}).

  \subsection{Closed-Book Contamination Filter}
  \label{sec:closed_book}

  To remove questions that can be answered from prior LLM knowledge alone, every one of the 1{,}604 questions is first attempted closed-book by three solver LLMs spanning two model families: \texttt{gpt-5.1}, \texttt{gpt-4.1-mini-2025-04-14}, and \texttt{gemini-2.5-flash}. Each closed-book attempt uses the same answering prompt as the main evaluation (Appendix~\ref{sec:appendix_prompts}, ``Answering Prompt'') with the context descriptor set to the \emph{Closed-book} variant: no case file or retrieved excerpt is supplied, and the solver is instructed to rely on prior knowledge and to abstain when the answer cannot be determined with confidence. Closed-book calls use the same temperature, retry budget, and structured-output schemas as the main evaluation. A question is flagged as priors-guessable when at least two of the three solvers commit to the correct answer; abstentions and wrong commits both count as non-correct, so the rule selects only items where multiple models actively assert the correct answer without context. This procedure flagged $190$ questions (11.8\%), leaving $1{,}414$ questions for all evaluations reported in \S\ref{sec:evaluation} and the remainder of this appendix.

  \subsection{Long-Context Baselines}

  The long-context architecture places the full case file (50K--100K tokens) directly in the LLM's prompt alongside the question; no chunking, retrieval, or compression is applied. We evaluate eight snapshot-pinned models: five closed-source models (\texttt{gpt-5.1}, \texttt{gpt-4o-mini-2024-07-18}, \texttt{gpt-4.1-mini-2025-04-14}, \texttt{gemini-2.5-pro}, \texttt{gemini-2.5-flash}) and three open-weight models (\texttt{llama-3.3-70b}, \texttt{qwen3-vl-30b}, \texttt{kimi-k2.5}). Each question is answered independently with the full case file in context; per-model concurrency is $5$.

  \subsection{Oracle}

  The Oracle replaces the narrated case file with the structured ground-truth representation from which it was generated, namely the provenance DAG of entities, timestamped events, evidence items, invalidations, and counterfactual dependencies described in \S\ref{sec:generation}. This structured representation is serialized into the prompt in place of the narrative text. Oracle is evaluated with two answering models, \texttt{gpt-5.1} and \texttt{gemini-2.5-flash}, the same models used for the RAG and memory architectures; this provides a retrieval-perfect upper bound under matched answering capacity.

  \subsection{RAG}

  All four RAG variants share a common backbone. The case file is chunked at log-entry boundaries (one chunk per timestamped entry, approximately $800$ chunks per case), embedded with \texttt{text-embedding-3-large}, and stored in a local ChromaDB collection keyed by case identifier with cosine similarity. At query time the question text, with MCQ option text appended for richer retrieval signal, is embedded with the same model. Retrieved chunks are concatenated and placed in the system prompt; the answering LLM sees only the retrieved chunks, not the full case file. Each variant is evaluated with two answering LLMs (\texttt{gpt-5.1} and \texttt{gemini-2.5-flash}), matching the Oracle setup; per-variant scores in Table~\ref{tab:main-results} are the unweighted mean across the two answering LLMs, with the per-LLM breakdown reported in Table~\ref{tab:per-llm-rag-memory}. The four variants differ only in the retrieval stage:

  \begin{itemize}[noitemsep, topsep=2pt, leftmargin=*]
      \item \textbf{plain dense}: top-$50$ dense retrieval, no reranking, no query expansion.
      \item \textbf{$+$rerank}: top-$50$ dense candidates are re-scored by the Cohere \texttt{rerank-v3.5} cross-encoder; the top-$50$ after reranking are passed to the LLM.
      \item \textbf{hybrid$+$rerank}: BM25 and dense each retrieve top-$50$, the two candidate sets are fused via Reciprocal Rank Fusion, and the fused list is Cohere-reranked to a final top-$50$.
      \item \textbf{hybrid$+$rerank$+$multi-query}: the question is rewritten into two paraphrases by \texttt{gpt-4o-mini} (three queries in total); each query retrieves top-$30$ from dense and BM25, the candidates are RRF-fused and Cohere-reranked to a final top-$50$.
  \end{itemize}

  \subsection{Mem0 and Mem0-Graph}

  Mem0 is a cloud memory service that extracts and indexes atomic facts from text. During ingestion, consecutive case-file entries are batched into approximately $4{,}096$-character blocks, matching the MemoryAgentBench reference chunk size, and submitted to Mem0 with day, time, entry type, and author as metadata. Mem0 extracts atomic facts, deduplicates them, and builds a searchable index. We wait $30$ seconds after ingestion to cover Mem0's documented asynchronous fact-extraction window. At query time we retrieve the top-$50$ most relevant memories, which are concatenated and passed to the answering LLM as paraphrased atomic facts rather than verbatim text. Mem0 is evaluated with two answering LLMs (\texttt{gpt-5.1} and \texttt{gemini-2.5-flash}); per-LLM scores are reported in Table~\ref{tab:per-llm-rag-memory}.

  \textbf{Mem0-Graph} uses the same backend with graph extraction enabled, which additionally extracts entity--relation triples during ingestion and returns them alongside retrieved memories. The two fields are appended in sequence into the answering LLM's context (memories first, then triples) with no joint reranking. We report it as a distinct system because the graph layer changes what is retrievable per question, even though the underlying service is shared with Mem0. Retrieval uses top-$30$ memories, and the ingestion settle interval is increased to $60$ seconds to cover Mem0's documented graph-construction window. Mem0-Graph is evaluated under the same two-LLM answering protocol as Mem0.

  \subsection{Hindsight}

  Hindsight is a self-hosted memory agent with a retain--recall--reflect pipeline. During setup we create a per-case memory bank with a mission statement describing the investigation domain and a forensic-analysis disposition profile. Ingestion uses Hindsight's batch interface, with each entry mapped from its in-case timestamp (e.g., Day~5, 14{:}30) to an absolute datetime so that Hindsight's temporal search strategies become available. We wait $60$ seconds after ingestion to cover Hindsight's documented background observation-consolidation window. At query time we use Hindsight's \emph{recall} mode (rather than \emph{reflect}), at the maximum recall budget and with raw source chunks included, returning both consolidated memories and up to $1{,}500$ tokens of source text. These are concatenated and passed to the answering LLM. Hindsight is evaluated with two answering LLMs (\texttt{gpt-5.1} and \texttt{gemini-2.5-flash}); per-LLM scores are reported in Table~\ref{tab:per-llm-rag-memory}.

  \subsection{Supermemory}

  Supermemory is a cloud document-memory store with semantic search. Each case-file entry is ingested under a per-case container tag, with deterministic per-entry identifiers to support idempotent re-ingestion. We wait $600$ seconds after ingestion, the upper bound of Supermemory's documented indexing window for documents of this size. At query time we query Supermemory's document-search endpoint with full-document return, server-side reranking, and query rewriting enabled, so that full source documents (rather than chunk previews) are returned under Supermemory's own reranking and query-rewriting layers. The top-$50$ documents are concatenated and passed to the answering LLM. Supermemory is evaluated with two answering LLMs (\texttt{gpt-5.1} and \texttt{gemini-2.5-flash}); per-LLM scores are reported in Table~\ref{tab:per-llm-rag-memory}.

  \subsection{Human Baseline}
  \label{sec:appendix_human}

  \textbf{Sample and protocol.} We sample 60 questions stratified across the six task categories (10 per category) and balanced across the three domains ($\sim$20 per domain). Three annotators independently attempted each question with access to a question-specific evidence packet (the answer-relevant provenance subgraph paired with its matching narrative excerpts) but no access to the structured skeleton, the full case file, or any oracle aid. The abstain option was available throughout. Each annotator spent approximately 9 hours total ($\sim$8.5--9 minutes per question). Final answers are determined by majority vote.

  \textbf{Per-task results.} Human Accuracy ranges from $90\%$ on Temporal Fact Retrieval to $37\%$ on Counterfactual; abstention rises from $3\%$ to $24\%$ and mean time per question rises from $3$ to $15$ minutes (Table~\ref{tab:human-pertask}).

  \begin{table}[!ht]
  \centering
  \small
  \setlength{\tabcolsep}{5pt}
  \renewcommand{\arraystretch}{0.95}
  \begin{tabular}{lrrrr}
  \toprule
  Task & Acc.\ & Score & Abst.\ & Time \\
  \midrule
  Temporal Fact Retrieval & 90\% & 0.88 &  3\% & 3 min \\
  Temporal Constraint     & 78\% & 0.72 &  6\% & 5 min \\
  Source Conflict         & 70\% & 0.63 &  8\% & 7 min \\
  Chain Reconstruction    & 58\% & 0.49 & 12\% & 9 min \\
  Cascade Propagation     & 45\% & 0.32 & 18\% & 12 min \\
  Counterfactual          & 37\% & 0.21 & 24\% & 15 min \\
  \bottomrule
  \end{tabular}
  \caption{Human baseline by task category (majority vote of three annotators per question).}
  \label{tab:human-pertask}
  \end{table}

  \textbf{Per-domain results.} Crime $65.0\%$, Finance $58.0\%$, and Medical $66.0\%$ Accuracy (Table~\ref{tab:human-perdomain}).

  \begin{table}[!ht]
  \centering
  \small
  \setlength{\tabcolsep}{8pt}
  \renewcommand{\arraystretch}{0.95}
  \begin{tabular}{lr}
  \toprule
  Domain & Accuracy \\
  \midrule
  Crime    & 65.0\% \\
  Finance  & 58.0\% \\
  Medical  & 66.0\% \\
  \bottomrule
  \end{tabular}
  \caption{Human baseline by domain.}
  \label{tab:human-perdomain}
  \end{table}

  \textbf{Comparison to Oracle.} The packet-equipped human baseline serves as a complementary upper bound to the Oracle solver, which receives the same answer-relevant evidence in structured form. Humans exceed Oracle Accuracy by $8.4$~pp ($63.0\%$ vs.\ $54.6\%$) at higher abstention ($11.8\%$ vs.\ $\sim$$5\%$), indicating that reasoning over presented evidence still leaves substantial headroom even when retrieval is controlled for. The per-task ordering aligns across humans, Oracle, and LLMs: Temporal Fact Retrieval is easiest and Counterfactual is hardest across all three.

  \subsection{Per-LLM and Per-Domain Breakdowns}

  Table~\ref{tab:per-llm-rag-memory} reports per-answering-LLM scores for the RAG and memory architectures (the family-mean rows in Table~\ref{tab:main-results}). Table~\ref{tab:by-domain} reports per-domain Score across all evaluated systems.

  \begin{table}[t]
  \centering
  \footnotesize
  \setlength{\tabcolsep}{8pt}
  \renewcommand{\arraystretch}{0.9}
  \begin{tabular}{lcc}
  \toprule
  System & \texttt{gpt-5.1} & \texttt{gemini-2.5-flash} \\
  \midrule
  \multicolumn{3}{l}{\textbf{RAG}}\\
  plain                       & 0.197 & 0.158 \\
  $+$rerank                   & 0.207 & 0.178 \\
  hybrid $+$ rerank           & 0.176 & 0.182 \\
  hybrid $+$ rerank $+$ MQ    & 0.161 & 0.184 \\
  \midrule
  \multicolumn{3}{l}{\textbf{Memory}}\\
  Mem0          & 0.106 & 0.118 \\
  Mem0-Graph    & 0.140 & 0.158 \\
  Supermemory   & 0.232 & 0.189 \\
  Hindsight     & 0.142 & 0.154 \\
  \bottomrule
  \end{tabular}
  \caption{Per-answering-LLM overall Score for RAG and Memory systems. The corresponding rows in Table~\ref{tab:main-results} are the unweighted mean of the two columns.}
  \label{tab:per-llm-rag-memory}
  \end{table}

  \begin{table*}[t]
  \centering
  \footnotesize
  \setlength{\tabcolsep}{7pt}
  \renewcommand{\arraystretch}{0.9}
  \begin{tabular}{lcccc}
  \toprule
  System & Crime & Medical & Finance & Overall \\
  \midrule
  \multicolumn{5}{l}{\textbf{Long-context}}\\
  gpt-5.1          & 0.304 & 0.303 & 0.253 & \textbf{0.287} \\
  gpt-4o-mini      & 0.188 & 0.246 & 0.273 & 0.232 \\
  gpt-4.1-mini     & 0.227 & 0.237 & 0.154 & 0.206 \\
  gemini-2.5-pro   & 0.435 & 0.193 & 0.127 & 0.265 \\
  gemini-2.5-flash & 0.305 & 0.254 & 0.200 & 0.256 \\
  kimi-k2.5        & 0.314 & 0.180 & 0.256 & 0.257 \\
  qwen3-vl-30b     & 0.280 & 0.244 & 0.159 & 0.230 \\
  llama-3.3-70b    & 0.176 & 0.161 & 0.237 & 0.192 \\
  \midrule
  \multicolumn{5}{l}{\textbf{Oracle}}\\
  gpt-5.1          & 0.561 & \textbf{0.726} & 0.655 & 0.638 \\
  gemini-2.5-flash & \textbf{0.689} & 0.600 & \textbf{0.657} & \textbf{0.654} \\
  \midrule
  \multicolumn{5}{l}{\textbf{RAG}}\\
  plain            & 0.263 & 0.112 & 0.133 & 0.178 \\
  +rerank          & 0.256 & 0.191 & 0.118 & 0.192 \\
  hybrid + rerank  & 0.278 & 0.129 & 0.106 & 0.179 \\
  hybrid + MQ      & 0.251 & 0.186 & 0.068 & 0.173 \\
  \midrule
  \multicolumn{5}{l}{\textbf{Memory}}\\
  Mem0             & 0.106 & 0.133 & 0.102 & 0.112 \\
  Mem0-Graph       & 0.176 & 0.149 & 0.116 & 0.149 \\
  Supermemory      & 0.299 & -0.020 & 0.150 & 0.211 \\
  Hindsight        & 0.157 & 0.162 & 0.125 & 0.148 \\
  \bottomrule
  \end{tabular}
  \caption{Per-domain scores across all evaluated systems.}
  \label{tab:by-domain}
  \end{table*}

\subsection{Per-task Family-Mean Visualization}

\begin{figure}[t]
\centering
% Family-aggregate per-task calibrated Score.
% Tasks ordered by Oracle - best-non-Oracle Score gap (descending).
% Each bar is the mean Score across systems within a family.
% Source: Table 3 in the main paper (post-G1-G5 corrections).
%
% Long-context mean over 8 LLMs.
% RAG mean over 4 retrieval variants.
% Memory mean over Mem0, Mem0-Graph, Supermemory, Hindsight.
% Oracle mean over gpt-5.1 + gemini-2.5-flash.

\definecolor{famOracle}{RGB}{35,75,115}    % deep navy
\definecolor{famLong}{RGB}{217,95,2}        % orange
\definecolor{famRAG}{RGB}{27,158,119}       % teal
\definecolor{famMem}{RGB}{117,112,179}      % purple

\begin{tikzpicture}
\begin{axis}[
    ybar,
    bar width=4.0pt,
    enlarge x limits=0.10,
    width=0.85\columnwidth,
    height=5.0cm,
    ymin=0, ymax=1.10,
    ytick={0,0.2,0.4,0.6,0.8,1.0},
    ylabel={Calibrated Score},
    ylabel style={font=\small},
    symbolic x coords={SrcConf,Chain,Cascade,Counter,TConstr,TFact},
    xtick=data,
    xticklabel style={font=\footnotesize, rotate=25, anchor=north east, xshift=2pt, yshift=1pt},
    yticklabel style={font=\small},
    ymajorgrids=true,
    major grid style={dashed,gray!25,line width=0.3pt},
    axis line style={line width=0.5pt},
    tick style={line width=0.4pt,black!60},
    legend style={
        font=\footnotesize,
        at={(0.5,1.03)},
        anchor=south,
        legend columns=2,
        column sep=0.4cm,
        row sep=0.05cm,
        draw=none,
        fill=none,
    },
    legend image code/.code={
        \draw[#1, draw=none] (0cm,-0.10cm) rectangle (0.30cm,0.10cm);
    },
    nodes near coords,
    nodes near coords style={
        font=\tiny,
        rotate=90,
        anchor=west,
        inner sep=1pt,
        /pgf/number format/.cd, fixed, fixed zerofill, precision=2,
    },
    nodes near coords align={vertical},
]
\addplot[fill=famOracle, draw=famOracle!80!black] coordinates {
    (SrcConf,0.948) (Chain,0.834) (Cascade,0.931) (Counter,0.479) (TConstr,0.337) (TFact,0.339)
};
\addplot[fill=famLong, draw=famLong!80!black] coordinates {
    (SrcConf,0.258) (Chain,0.223) (Cascade,0.305) (Counter,0.224) (TConstr,0.134) (TFact,0.314)
};
\addplot[fill=famRAG, draw=famRAG!80!black] coordinates {
    (SrcConf,0.089) (Chain,0.082) (Cascade,0.283) (Counter,0.181) (TConstr,0.134) (TFact,0.378)
};
\addplot[fill=famMem, draw=famMem!80!black] coordinates {
    (SrcConf,0.067) (Chain,0.124) (Cascade,0.574) (Counter,0.064) (TConstr,0.140) (TFact,0.147)
};
\legend{Oracle, Long-context, RAG, Memory}
\end{axis}
\end{tikzpicture}
\caption{Mean calibrated Score by family across the six task categories.}
\label{fig:per-task-bars}
\end{figure}
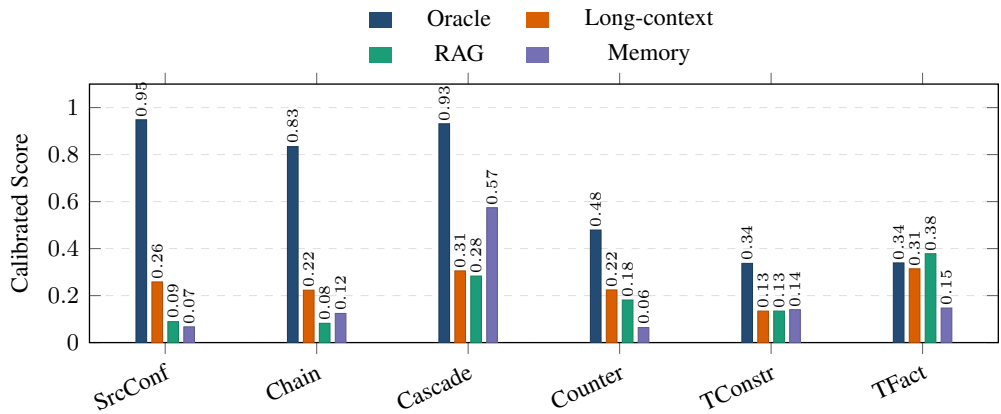

Figure~\ref{fig:per-task-bars} aggregates calibrated Score by family across the six task categories, complementing the per-system breakdown in Table~\ref{tab:main-results}. The Oracle lead is strongly task-specific: 50--70 pp on the three structured tasks (Cascade, Chain Reconstruction, Source Conflict), 15--35 pp on Counterfactual and Temporal Constraint, and disappears on Temporal Fact Retrieval where RAG and long-context overtake the structured representation. Memory averages $0.57$ on Cascade but does not exceed $0.15$ on any other task: Chain Reconstruction and Source Conflict require inter-fact edges that document-level and atomic-fact stores compress poorly.

  \clearpage
  \section{Example Questions}
  \label{sec:appendix_examples}

  Below we present one MCQ-single example per task category drawn from the evaluated case file. Each question has six options (A--F); the correct answer is marked with $\checkmark$. Options are designed as targeted distractors that exploit specific failure modes (noted in parentheses where relevant).

  \subsection{Chain Reconstruction}

  \textit{Which statement correctly summarizes the evidence trail ``Tracking Marcus Webb's movements on the day of the murder''?}

  \begin{enumerate}[label=\Alph*), noitemsep, topsep=2pt, font=\small]
    \item The correct sequence is: Marcus Webb admits to purchasing cyanide \ldots\ $\to$ Marcus Webb's card swiped into university library at 5:05 PM. \hfill {\small(reversed order)}
    \item The trail involves the first and last steps but omits the intermediate steps: library swipe-out, witness sighting, traffic camera, \ldots \hfill {\small(incomplete chain)}
    \item The key steps are the initial discovery and the final confirmation; the middle steps are procedural filler. \hfill {\small(minimizes chain)}
    \item This trail includes ``Emails containing threats found on victim's computer'' as a key step. \hfill {\small(wrong chain)}
    \item The most recent finding is the only step needed to establish the trail. \hfill {\small(single-hop shortcut)}
    \item[$\checkmark$ F)] Step 1: Library swipe-in (5:05 PM) $\to$ Step 2: Swipe-out (6:42 PM) $\to$ Step 3: Witness sighting (6:45 PM) $\to$ Step 4: Traffic camera (7:15 PM) $\to$ Step 5: Phone disconnected (7:00--9:00 PM) $\to$ Step 6: Coffee mug found $\to$ Step 7: Fingerprint match $\to$ Step 8: Cyanide receipt $\to$ Step 9: Credit card link $\to$ Step 10: Confession.
  \end{enumerate}

  \noindent\textit{Note: This question has chain length 10, requiring the model to locate and correctly order all 10 hops scattered across the 100K-token case file.}

  \subsection{Cascade Propagation}

  \textit{After the contamination of the coffee mug with residue found at the victim's desk in Office 304 (detective accidentally smudged fingerprints while collecting evidence), which conclusion is invalidated?}

  \begin{enumerate}[label=\Alph*), noitemsep, topsep=2pt, font=\small]
    \item The contamination occurred after all forensic analysis was completed, so no results are affected. \hfill {\small(ignores invalidation)}
    \item[$\checkmark$ B)] Fingerprint matches to Marcus Webb.
    \setcounter{enumi}{2}
    \item All forensic results from the coffee mug remain fully valid despite the contamination. \hfill {\small(ignores invalidation)}
    \item Autopsy findings (independent of physical evidence). \hfill {\small(over-scoping)}
    \item Both fingerprint matches and cyanide detection are compromised; all evidence from the mug should be excluded entirely. \hfill {\small(over-scoping)}
    \item All other forensic results processed separately from the coffee mug. \hfill {\small(wrong target)}
  \end{enumerate}

  \noindent\textit{Note: The model must recognize that cyanide detection (a chemical test) is independent of fingerprint contamination and survives the invalidation. Option E is the primary trap, testing whether the model collapses all downstream conclusions without checking for independent support.}

  \subsection{Source Conflict Resolution}

  \textit{Which statement correctly describes the resolution of the conflict between Laura Kim and David Nguyen?}

  \begin{enumerate}[label=\Alph*), noitemsep, topsep=2pt, font=\small]
    \item Laura Kim's statement was recorded first, so it should be considered more reliable. \hfill {\small(recency bias)}
    \item Both claims can be true simultaneously. \hfill {\small(false reconciliation)}
    \item[$\checkmark$ C)] Marcus Webb left the library as witnessed by Laura Kim, heading towards the clinic.
    \setcounter{enumi}{3}
    \item David Nguyen's professional role makes their observation more credible. \hfill {\small(authority bias)}
    \item Both witnesses are equally credible with no way to determine which is correct. \hfill {\small(false equivalence)}
    \item The corroborating evidence (traffic camera at 7:15 PM) actually supports David Nguyen's account. \hfill {\small(misattributed corroboration)}
  \end{enumerate}

  \noindent\textit{Note: The traffic camera sighting near the clinic at 7:15 PM corroborates Kim's account (Webb left the library toward the clinic), not Nguyen's claim that Webb was at a caf\'{e}.}

  \subsection{Counterfactual Reasoning}

  \textit{If the contamination of the coffee mug with residue had been discovered on Day 9 instead of Day 11, would Detective Ruiz's notes on Day 9 still have occurred?}

  \begin{enumerate}[label=\Alph*), noitemsep, topsep=2pt, font=\small]
    \item This investigative action would have taken place earlier. \hfill {\small(temporal confusion)}
    \item Cannot be determined from the available information. \hfill {\small(false uncertainty)}
    \item This action would have taken place but led to different conclusions. \hfill {\small(partial counterfactual)}
    \item[$\checkmark$ D)] No, this step would not have occurred.
    \setcounter{enumi}{4}
    \item This finding would still have occurred, just at a later point. \hfill {\small(temporal shift)}
    \item Yes, this step would still have occurred. \hfill {\small(ignores dependency)}
  \end{enumerate}

  \noindent\textit{Note: The model must determine that Detective Ruiz's Day 9 notes depended on the coffee mug evidence, so discovering the contamination on Day 9 (before the notes were written) would have prevented them.}

  \subsection{Temporal Constraint Satisfaction}

  \textit{Which suspect lacks continuous surveillance or cell tower coverage during the time of the crime?}

  \begin{enumerate}[label=\Alph*), noitemsep, topsep=2pt, font=\small]
    \item Nina Patel.
    \item[$\checkmark$ B)] Marcus Webb.
    \setcounter{enumi}{2}
    \item Lydia Tran.
    \item Samuel Lee.
    \item Robert Harris.
    \item Eleanor Dawson.
  \end{enumerate}

  \noindent\textit{Note: Marcus Webb's phone was disconnected from the network between 7:00 PM and 9:00 PM on Day 1, creating a gap in cell tower coverage during the crime window. All other suspects have continuous coverage from at least one data stream.}

  \subsection{Temporal Fact Retrieval (Baseline)}

  \textit{Which happened first: a cell tower ping recorded at Hillcrest Suites or Detective Ruiz's alibi verification for Samuel Lee?}

  \begin{enumerate}[label=\Alph*), noitemsep, topsep=2pt, font=\small]
    \item A cell tower ping recorded for Lydia Tran. \hfill {\small(wrong person)}
    \item Officer Davis's alibi verification for Samuel Lee. \hfill {\small(wrong detective)}
    \item Detective Kowalski's witness statement about Nancy Green. \hfill {\small(wrong event type)}
    \item Officer Chen's surveillance report involving Jonathan Cruz. \hfill {\small(wrong event type)}
    \item Detective Ruiz's alibi verification for Samuel Lee. \hfill {\small(wrong temporal order)}
    \item[$\checkmark$ F)] A cell tower ping recorded at Hillcrest Suites.
  \end{enumerate}

  \noindent\textit{Note: The Hillcrest Suites cell tower ping occurred on Day 1 (18:17), while Detective Ruiz's alibi verification occurred on Day 4 (07:46). This is a straightforward temporal ordering question requiring no multi-hop reasoning.}

  \end{document}